\documentclass[A4]{IEEEtran}
\IEEEoverridecommandlockouts
\usepackage{cite}
\usepackage{amsmath,amssymb,amsfonts}
\usepackage{amssymb}
\usepackage{breqn}
\usepackage{cleveref}
\usepackage{booktabs}
\usepackage{cases}
\usepackage{graphicx}
\usepackage{multirow}
\usepackage{mathrsfs}
\usepackage{textcomp}
\usepackage{xcolor}
\usepackage{subfigure}
\usepackage{bm}
\usepackage{graphicx}
\usepackage{array}
\usepackage{algorithm}
\usepackage{algorithmicx}
\usepackage{algpseudocode}

\newcommand{\ignore}[1]{}
\def\BibTeX{{\rm B\kern-.05em{\sc i\kern-.025em b}\kern-.08em
    T\kern-.1667em\lower.7ex\hbox{E}\kern-.125emX}}

\begin{document} 

\title{Neuro-Symbolic Multitasking: A Unified Framework for Discovering Generalizable Solutions to PDE Families}

\author{
    Yipeng~Huang,~\IEEEmembership{}
    Dejun~Xu,~\IEEEmembership{}
    Zexin~Lin,~\IEEEmembership{}
    Zhenzhong~Wang,~\IEEEmembership{Member,~IEEE,}
    Min Jiang,~\IEEEmembership{Senior~Member,~IEEE}

	\IEEEcompsocitemizethanks{\IEEEcompsocthanksitem
 
Yipeng Huang, Dejun Xu, Zexin Lin, Min Jiang and Zhenzhong Wang are with the Department of Artificial Intelligence, Key Laboratory of Digital Protection and Intelligent Processing of Intangible Cultural Heritage of Fujian and Taiwan, Ministry of Culture and Tourism, School of Informatics, Xiamen University, and Key Laboratory of Multimedia Trusted Perception and Efficient Computing, Ministry of Education of China, Xiamen University, Xiamen 361005, Fujian, China  \textit{(Corresponding authors: Zhenzhong Wang and Min Jiang, e-mail: zhenzhongwang@xmu.edu.cn and minjiang@xmu.edu.cn)}.


	}	
}

\bibliographystyle{IEEEtran}
\maketitle
\begin{abstract}

Solving Partial Differential Equations (PDEs) is fundamental to numerous scientific and engineering disciplines. A common challenge arises from solving the PDE families, which are characterized by sharing an identical mathematical structure but varying in specific parameters. 
Traditional numerical methods, such as the finite element method, need to independently solve each instance within a PDE family, which incurs massive computational cost. On the other hand, while recent advancements in machine learning PDE solvers offer impressive computational speed and accuracy, their inherent ``black-box" nature presents a considerable limitation. These methods primarily yield numerical approximations, thereby lacking the crucial interpretability provided by analytical expressions, which are essential for deeper scientific insight. 
To address these limitations, we propose a neuro-assisted multitasking symbolic PDE solver framework for PDE family solving, dubbed NMIPS. 
In particular, we employ multifactorial optimization to simultaneously discover the analytical solutions of PDEs. To enhance computational efficiency, we devise an affine transfer method by transferring learned mathematical structures among PDEs in a family, avoiding solving each PDE from scratch. Experimental results across multiple cases demonstrate promising improvements over existing baselines, achieving up to a $\sim$35.7\% increase in accuracy while providing interpretable analytical solutions. 

\end{abstract}


\section{INTRODUCTION}

Solving Partial Differential Equations (PDEs) lies at the core of numerous scientific and engineering disciplines, providing the mathematical language to describe phenomena ranging from fluid dynamics to quantum mechanics~\cite{bergen2019machine,kovachki2023neural,oldenburg2022geometry,amaral2025quantum}. Within this broad field, a particularly common and challenging scenario involves families of PDEs: these are sets of equations that share an identical underlying mathematical structure but vary in specific parameters. For instance, consider the need to simulate heat transfer in a new alloy across a range of operational temperatures, or to model the flow of a chemical mixture under various concentrations. In such cases, the fundamental governing PDE remains the same, but the specific material properties or environmental conditions, represented by parameters, change~\cite{kovachki2023neural}.

Early methodologies for solving PDE families have heavily relied on numerical techniques, such as the Finite Element Method (FEM)~\cite{Tekkaya2014} or Finite Volume Method (FVM)~\cite{eymard2000finite}. These approaches typically operate by discretizing the domain to iteratively solve each individual PDE instance. Despite their robustness and established theoretical foundations, which offer rigorous guarantees regarding stability and convergence, these methods suffer from prohibitive computational costs. This limitation renders them ill-suited for high-throughput scenarios or real-time applications, as every parameter variation necessitates a complete re-computation from scratch.


More recently, the landscape of PDE solving has been significantly transformed by advancements in machine learning-based solvers, creating distinct paradigms. Among these architectures, Physics-Informed Neural Networks (PINNs)~\cite{cai2021physics, oldenburg2022geometry}, depicted in Fig.~\ref{fig:motivation} (a), directly embed physical laws into the network to enforce consistency with the governing equations. However, PINNs remain instance-specific, requiring retraining for new conditions. To overcome this, Neural Operators (NOs)~\cite{kovachki2023neural, li2021fourier, cao2024laplace} shown in Fig.~\ref{fig:motivation} (b) have emerged as generalizable alternatives. These methods leverage universal approximation capabilities to learn mappings from input parameters directly to PDE solutions. Consequently, NOs achieve substantial computational efficiency; once trained, they utilize their generalization capabilities to perform direct inference across varying parameter configurations without iterative re-optimization.


\begin{figure}[t]
    \centering
    \includegraphics[width=0.49\textwidth]{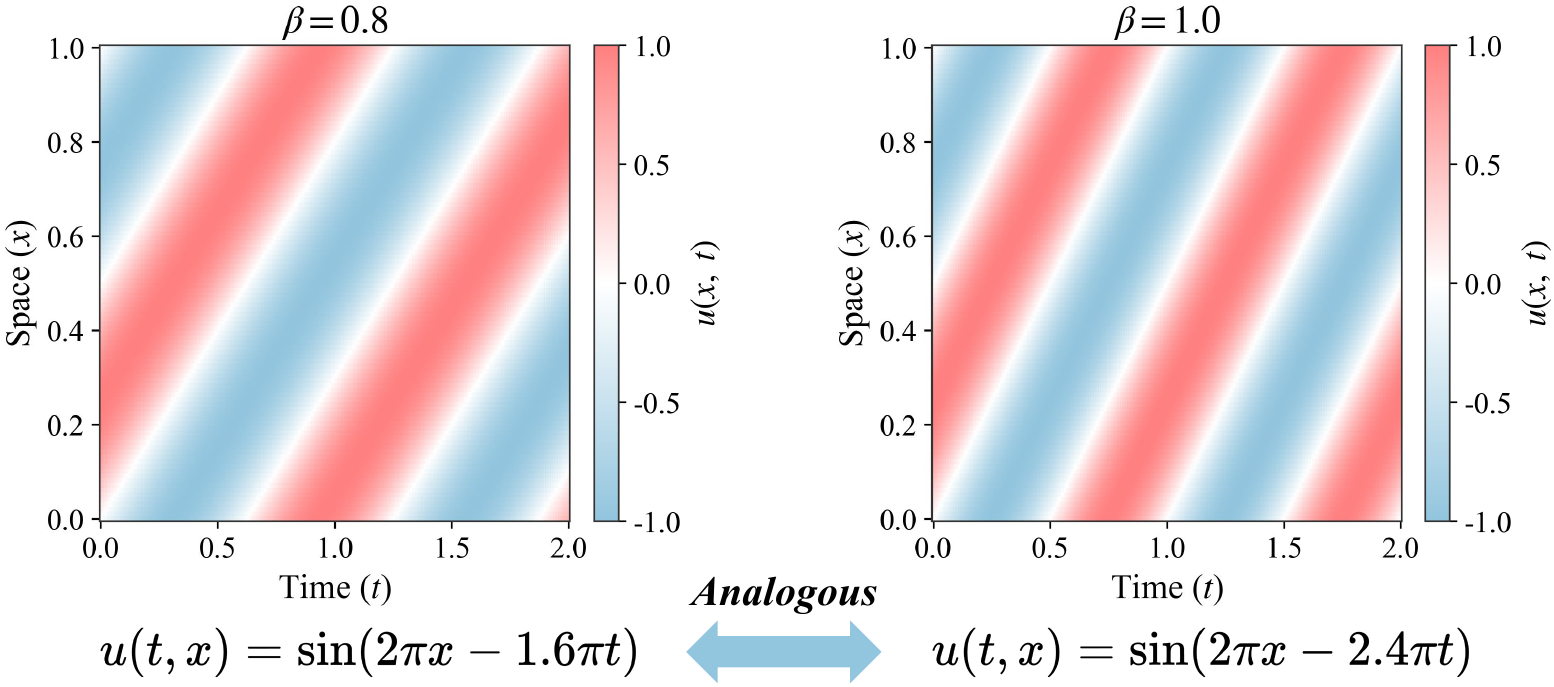}
    \caption{Numerical solutions’ landscapes for PDEs with different parameters reveal that equations belonging to the same family demonstrate analogous analytical solutions.} 
    \label{fig:introduction}
\end{figure}

While these machine learning-based PDE solvers offer substantial computational speed and accuracy, their inherent ``black-box" nature presents a considerable limitation. These methods only provide numerical approximations, thereby lacking the crucial interpretability inherent to analytical expressions~\cite{cao2025interpretable, cao2024interpretable}. Such interpretability is essential for deeper scientific insight and broader applicability. For instance, in analyzing complex fluid dynamics or reaction-diffusion systems, obtaining explicit symbolic expressions (such as specific convection terms or diffusion coefficients) allows researchers to directly identify the physical mechanisms and conservation laws governing the system dynamics~\cite{brunton2016discovering}.

\begin{figure*}[t]
    \centering
    \includegraphics[width=1.0\textwidth]{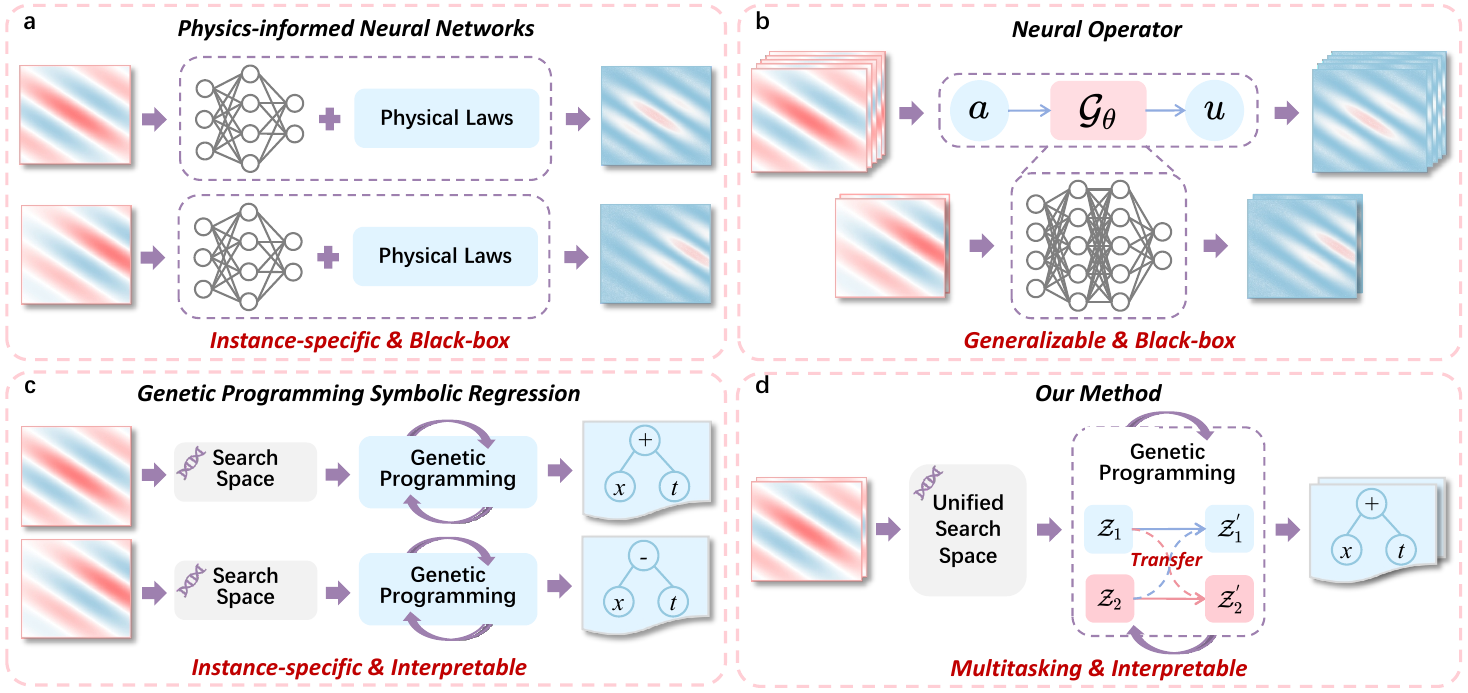}
    \caption{Schematic comparison of different paradigms for solving PDEs. (a) Physics-informed Neural Networks embed physical laws by constraining the network training with PDE residuals. Despite integrating physics, they yield uninterpretable numerical approximations and require retraining for each instance, operating as instance-specific black-box models. (b) Neural Operator achieves generalizability by learning mappings between function spaces; however, they essentially remain black-box models that output numerical solutions without analytical forms. (c) Genetic Programming Symbolic Regression offers interpretability by searching for analytical expressions, yet it is constrained to solving one specific problem at a time. (d) Our method constructs a unified search space and leverages a transfer mechanism between tasks, realizing a paradigm that is both multitasking capable and interpretable.} 
    \label{fig:motivation}
\end{figure*}

To bridge this gap, symbolic regression (SR) methods offer a powerful alternative capable of discovering closed-form mathematical formulas, providing highly interpretable analytical solutions for PDEs. However, a key limitation of existing symbolic regression techniques, represented by Genetic Programming Symbolic Regression (GPSR)~\cite{cao2023genetic} in Fig.~\ref{fig:motivation} (c), lies in their singular focus; they are incapable of generalizing across cases, treating only one PDE instance at a time. This makes them inefficient for solving entire PDE families, as each parameter variation necessitates an independent and time-consuming discovery process.


Crucially, instances within a PDE family share an identical mathematical skeleton. This is empirically illustrated in Fig.~\ref{fig:introduction}, where the spatiotemporal landscapes of the 1D Advection equation are compared under different coefficients ($\beta=0.8$ and $\beta=1.0$). Despite the parametric variation, the numerical solutions exhibit similar wave propagation patterns, and their corresponding analytical expressions retain a consistent symbolic structure, differing only in specific scalar coefficients. This implies that their solution landscapes exhibit intrinsic structural correlations~\cite{quarteroni2015reduced, penwarden2023metalearning}. Explicitly leveraging these shared characteristics could bypass redundant discovery processes, accelerating the discovery of related equations.

Keeping this in mind, we propose a novel Multitasking Symbolic Regression framework (i.e., Fig.~\ref{fig:motivation} (d)) to simultaneously solve multiple PDEs within a given family, moving beyond the one-off nature of traditional symbolic regression. The proposed multitasking symbolic regression framework is devised based on the multifactorial optimization strategy\cite{gupta2015multifactorial}.
In particular, to promote the process of reusing shared knowledge among instances within a PDE family, we leverage an affine transfer method, which exploits and transfers the shared mathematical structural features that inherently exist between PDEs belonging to the same family, enabling more rapid and effective discovery of their analytical solutions. The main contributions are follows:

\begin{enumerate}

\item We introduce the first framework capable of simultaneously solving entire families of PDEs to yield interpretable analytical solutions. Unlike existing methods that address individual PDE instances, our approach leverages the inherent structural similarities within a PDE family, offering a paradigm shift towards efficient and comprehensive solution discovery.

\item We employ a multifactorial optimization strategy to technically solve the PDE families simultaneously. By mapping distinct PDE instances into a unified optimization landscape, this mechanism enables a single population to concurrently evolve analytical solutions for multiple equations.

\item To promote the knowledge transfer, we integrate a novel affine transfer method to exploit the similarity between the instances within a PDE family. This technique transfers learned mathematical structural features between related PDEs, dramatically accelerating the process of finding analytical solutions compared to learning each PDE from scratch.

\item Experimental results across multiple cases demonstrate significant improvements over existing baselines, achieving up to a $\sim$35.7\% increase in accuracy while achieving promising efficiency. Furthermore, our proposed method showcased superior generalization capabilities in noisy data settings.

\end{enumerate}

The rest of this paper is organized as follows. Section~\ref{pre} briefly presents the basic concepts of PDEs and reviews existing studies. Section~\ref{method} details the proposed approach. In Section~\ref{experiment}, the empirical results of the proposed algorithm and various state-of-the-art algorithms are presented. Finally, conclusions are drawn in Section~\ref{conclusion}.

\section{PRELIMINARIES}
\label{pre}

\subsection{PDE Families}

PDEs serve as the foundation for describing physical phenomena across science and engineering. However, practical applications often focus on a PDE family rather than a single equation \cite{pdebench}. A PDE family consists of equations that share an identical underlying mathematical structure but differ in specific physical parameters, such as viscosity or diffusion coefficients.
In such families, the fundamental behavior of the system remains consistent, while the specific properties vary with the parameter settings.

Formally, a PDE family is defined by a general functional form $G$ that explicitly includes a parameter vector $\mathbf{p}$. A PDE belonging to such a family is expressed as:
\begin{equation}
G(x_1, \dots, x_n, u, \nabla u, \dots, \nabla^k u; \mathbf{p}) = 0.
\end{equation}
Here, $G$ outlines the relationships between the independent variables, the unknown function, and its derivatives, while each unique parameter combination in $\mathbf{p}$ specifies a distinct instance within the family. Therefore, solving a PDE family involves finding solutions $u(\mathbf{x}; \mathbf{p})$ across a specified range of parameter values. For a comprehensive mathematical background on general PDEs and illustrative examples, please refer to the supplemental material (see Section I).

\subsection{Related Work}

\textbf{a) Traditional Numerical Methods.} These established techniques primarily rely on domain discretization. The Finite Difference Method (FDM) approximates derivatives using finite difference formulas \cite{robertsson2015finite, wahyudi2021study,ullah2021entropy\ignore{,thomas2013numerical}}. Similarly, the FEM and FVM discretize the problem domain into elements or control volumes, utilizing polynomial basis functions to approximate solutions \cite{\ignore{polycarpou2022introduction, szabo2021finite, }marzokExtendedFiniteVolume2025, muhammad2021finite, haider2022dynamics}. In contrast, spectral methods employ global basis functions, such as Fourier series or Chebyshev polynomials, to approximate solutions across the entire domain \cite{bernardi1997spectral, burns2020dedalus, meuris2023machine}. While robust, the optimal choice among these paradigms depends heavily on the specific PDE characteristics, and they often face computational bottlenecks in high-resolution or multi-query scenarios.

\textbf{b) Deep Learning-based PDE Solvers.} To address computational limitations, deep learning methodologies have emerged as promising alternatives. Finite-dimensional operators typically employ deep convolutional neural networks to parameterize the solution, though they often suffer from mesh dependency \cite{kag2022condensing, nikolopoulos2022non}. PINNs integrate observational data with the governing PDE by embedding the PDE itself into the neural network's loss function, leveraging automatic differentiation for gradient computations \cite{karniadakis2021physics,buhendwa2021inferring\ignore{,mao2020physics},  yu2022gradient}. NOs, on the other hand, are designed to approximate the mapping from a PDE to its solution directly, yielding fast and discretization-invariant solvers \cite{mionet,codano, wei2023super, lu2021learning}. These methods have primarily been applied to solve individual PDEs. However, recent work by Cai \textit{et al}. \cite{cai2021deepm} extended the neural operator DeepONet \cite{lu2019deeponet} to address systems of PDEs.

\begin{figure*}[t]
    \centering
    \includegraphics[width=1.0\textwidth]{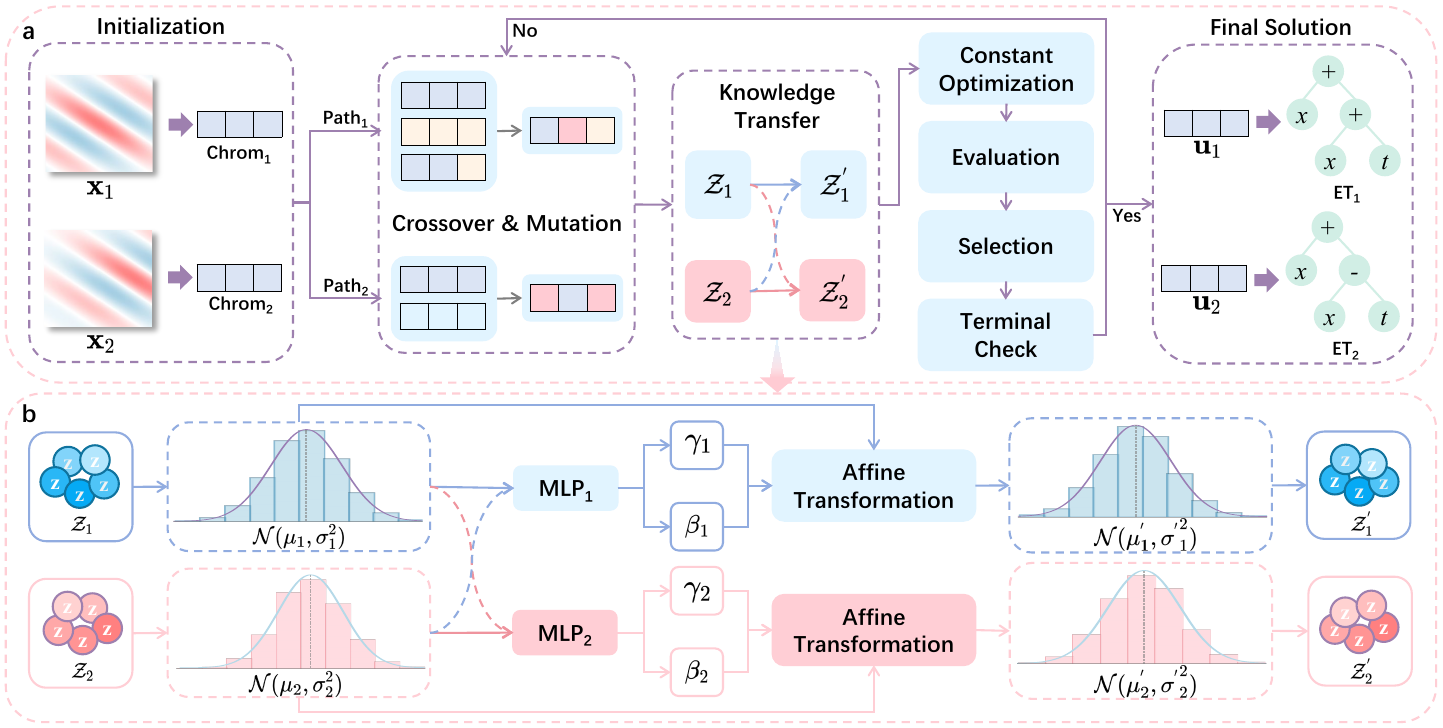}
    \caption{The overall framework of the proposed NMIPS. (a) Multifactorial PDE solving framework: This part constructs a unified encoding space to execute operations such as crossover, mutation, and selection. It embeds a knowledge transfer module to enable the efficient solving of multiple tasks within a PDE family. (b) Affine Transformation-based Knowledge Transfer: This mechanism extracts statistics of each population and uses multi-layer perceptrons to learn scaling and shifting parameters. It performs an affine transformation across populations to achieve efficient knowledge sharing between tasks.} 
    \label{fig:framework}
\end{figure*}

While these numerical and deep learning approaches provide effective approximations, the interpretability of their obtained solutions often remains a significant limitation. They typically function as black-box models or numerical approximations, failing to reveal the underlying analytical structures governing the physical systems. This limitation highlights the need for methods capable of discovering explicit analytical solutions, a challenge addressed by symbolic regression.



\textbf{c) Symbolic regression.} SR distinguishes itself from traditional solvers by searching for explicit mathematical expressions that describe observed data. In recent years, deep learning methods have made significant progress in this field. Unlike traditional discrete evolutionary search, these approaches leverage differentiable optimization and neural network guidance to improve search efficiency and robustness. 
For instance, Petersen \textit{et al}. proposed a deep symbolic regression method based on policy gradients, which generates symbolic expressions through risk-seeking strategies and effectively recovers complex functional relationships \cite{petersen2021deep}. Subsequently, Udrescu and Tegmark introduced the AI Feynman framework, which combines neural network approximations with physics-inspired decomposition strategies \cite{udrescu2020ai}. Moreover, end-to-end transformer models have enabled efficient search for high-dimensional expressions \cite{kamienny2022end}, and multi-modal fusion methods have further enhanced recoverability \cite{li2025mmsr}. However, despite their computational efficiency, these deep learning approaches often fall short in interpretability and precise control over symbolic structures.


In contrast, GPSR naturally offers advantages in interpretability and operator flexibility, making it particularly suitable for discovering analytical solutions to PDEs. GP can directly generate analytical expressions and explore complex structures through evolutionary operators. In recent years, a series of improvements has further enhanced the performance and generalization of GP. For instance, Huang \textit{et al}. introduced semantic linear genetic programming to accelerate convergence \cite{huang2022semantic}, while Bartlett \textit{et al}. developed an exhaustive framework providing theoretical guarantees for formula recovery in small-scale problems \cite{bartlett2023exhaustive}. Furthermore, to enhance search diversity and handle multi-objective complexities, Zhong \textit{et al}. proposed multiform GP frameworks \cite{zhong2025multiform}, while Wang \textit{et al}. integrated reinforcement learning with GP to achieve multi-task semantic optimization \cite{wang2025multi}. 

Specific to the domain of PDEs, researchers have increasingly leveraged these capabilities to solve physical equations. Oh \textit{et al}. used GP to search for symbolic functions satisfying boundary conditions, demonstrating that with judicious search space design, GP can discover analytical solutions for low-dimensional PDEs \cite{oh2023genetic}. To further ensure the parsimony and physical meaningfulness of the discovered equations, Cao \textit{et al}. introduced a simplification-pruning operator, yielding highly accurate and concise symbolic expressions \cite{cao2024interpretable}. Most recently, Cao \textit{et al}. further extended GP to multi-physics systems via T-NNGP, a hybrid algorithm combining numerical methods with deep learning for efficient solving \cite{cao2025interpretable}.

Nevertheless, despite these advancements, applying GP directly to the solution of PDEs remains challenging. On one hand, constraints inherent in PDEs involve higher-order derivatives and boundary conditions, whereas GP fitness functions typically rely solely on pointwise fitting errors. On the other hand, the high-dimensional spatiotemporal nature of PDEs makes traditional GP search prone to the curse of dimensionality. Most critically, however, a prevailing challenge remains: current methods generally struggle to simultaneously obtain analytical solutions for entire families of PDEs. They typically solve for a single instance with fixed parameters, lacking the generalization capability to handle parametric variations found in PDE families. Therefore, effectively applying GP to PDEs requires not only the integration of automatic differentiation and physical constraints but also mechanisms to support multi-task or generalizable learning, which is precisely the core problem addressed in this work.


\section{Methodology}
\label{method}


\subsection{Multifactorial Optimization for Multi-Task PDE Solving}

Assume a PDE family consists of $K$ PDE-solving tasks, $\mathcal{T}_i:\{\mathcal{L}_i(\mathbf{u}(\mathbf{x}_i)) = 0\}_{i=1}^K$, with different parameters, to be solved concurrently. Here, $\{\mathcal{L}_i\}_{i=1}^n$is a set of differential operators, {$\mathbf{u}(\mathbf{x})$} is a vector of unknown functions, and {$\mathbf{x} = (x, t)$} is a vector of spatial-temporal variables. The primary objective is to obtain a set of analytical solutions {$\{\mathbf{u}(\mathbf{x}_1), \mathbf{u}(\mathbf{x}_2), \dots, \mathbf{u}(\mathbf{x}_K)\}$}within a single optimization process to satisfy {$ \mathcal{L}_i(\mathbf{u}(\mathbf{x}_i)) = 0, \ i = 1, 2, \dots, K$}.
To achieve this, we propose NMIPS, a method that leverages multifactorial optimization to simultaneously tackle these multiple PDE tasks. The overall architectural framework is visually depicted in Fig.~\ref{fig:framework}, and the detailed algorithmic procedure is outlined in Algorithm~\ref{alg: NMIPS}.
In the following sections, we will introduce the various components of this multifactorial optimization formulation, including its chromosome structure, evolutionary operations, and population evaluation and selection.

\begin{figure}[t]
    \centering
    \includegraphics[scale=0.65]{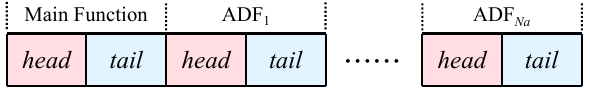}
    \caption{Structure of the C-ADF.} 
    \label{fig3}
\end{figure}

\textbf{a) Chromosome Design.} To represent the analytical solutions of PDEs, we employ expression trees (ETs) to encode these mathematical formulas. Structurally, ETs consist of internal function nodes and leaf nodes. Function nodes denote operations like addition ($+$), sine ($\sin$), etc., while leaf nodes represent terminals such as variables (e.g., $x, t$) and constants. Specifically, our chromosome structure is based on the gene expression representation with automatically defined functions (C-ADF)~\cite{andida2001gene, 8419217}. As illustrated in Fig.~\ref{fig3}, a C-ADF chromosome comprises one main function and multiple ADFs. The main function yields the final output of the solution, while the ADFs serve as reusable sub-functions called by the main function or other ADFs.

\begin{algorithm}[t]
    \small
    \caption{NMIPS}
    \label{alg: NMIPS}
    \begin{algorithmic}[1]
        \renewcommand{\algorithmicrequire}{\textbf{Input:}}
        \renewcommand{\algorithmicensure}{\textbf{Output:}}
        
        \Require \parbox[t]{0.94\linewidth}{\strut Population size $N$, Transfer interval $G_{tf}$, Random mating probability $rmp$.\strut}
        \Ensure The best analytical solutions for $K$ PDE tasks.
        \State Construct a unified chromosome encoding space compatible with all PDE tasks;
        \State Initialize a unified population within the space and calculate factorial costs with constant optimization;
        \State Assign the factorial rank, skill factor, and scalar fitness values for all individuals;
        
        \State Set generation counter $g=1$;
        \While{stopping conditions are not satisfied}
        
            \For{$i = 1$ to $N$}
                \If{$\text{rand}() < rmp$}
                    \State \parbox[t]{0.825\linewidth}{\strut Generate offspring $\mathbf{u}_i$ by one-point crossover and uniform mutation;\strut} 
                \Else
                    \State \parbox[t]{0.826\linewidth}{\strut Generate offspring $\mathbf{u}_i$ by DE-based mutation and crossover;\strut}
                \EndIf
                \State \parbox[t]{0.88\linewidth}{\strut Evaluate factorial cost of $\mathbf{u}_i$ with constant optimization based on its skill factor;\strut}
            \EndFor
            
            \State Perform one-to-one selection to update the population;
            \State Update scalar fitness for all individuals;
            
            \If{$g \mod G_{tf} == 0$}
                \State Perform \textbf{knowledge transfer} (see Algorithm~\ref{alg: KnowledgeTransfer});
            \EndIf
    
            \State Update best solutions for each task;
            \State $g = g + 1$;
        \EndWhile
        \State \textbf{return} The best analytical solutions found for all $K$ tasks;
    \end{algorithmic}
\end{algorithm}

\begin{figure}[t]
    \centering
    \includegraphics[scale=0.65]{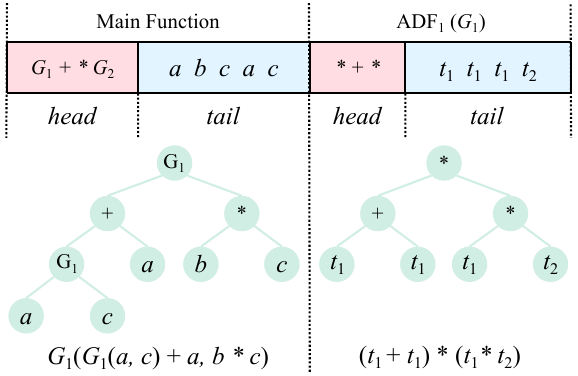}
    \caption{Example of a chromosome includes one main function and one ADF.} 
    \label{fig4}
\end{figure}

Both the main function and ADFs are represented by the Karva expression \cite{andida2001gene}. A Karva expression describes a solution using a fixed-length string composed of functions and terminals, which can be deterministically converted into an ET using a breadth-first traversal method. Each Karva expression is divided into two parts: a \emph{head} and a \emph{tail}. The head may contain both function and terminal symbols, whereas the tail consists only of terminals. To ensure that each chromosome can be converted into a valid ET, the lengths of the head ($h$) and tail ($l$) are constrained by the following relationship:
\begin{equation}\label{eq:karva_constraint}
l = h \cdot ({v} - 1) + 1,
\end{equation}
where {$v$} represents the maximum number of arguments among all functions in the defined function set. A specific example of a chromosome featuring one main function and one ADF, along with their decoded expressions, is provided in Fig.~\ref{fig4}. As observed from this example, a final solution can be decoded as $2 \cdot (2 \cdot a^2 \cdot c + a)^2 \cdot b \cdot c$.


To adapt C-ADF to a gene expression representation for multitasking, an extension to its encoding scheme is essential. A key challenge is that different tasks across domains may possess unique sets of functions and terminals. To address this, we employ integers to represent both functions and terminals. This approach allows a single integer value in the chromosome to represent various symbols specific to different tasks.

Specifically, suppose there are $K$ tasks to be solved. Let $F_i$ and $T_i$ denote the function set and terminal set of the $i$-th task, respectively. To ensure proper conversion of any chromosome to a valid ET, the lengths of the head ($h$) and tail ($l$) are governed by an adapted constraint:
\begin{equation}\label{eq:multitask_karva_constraint}
l = h \cdot \left( \max_{a \in F_1 \cup F_2 \cup \dots \cup F_K} \xi(a) - 1 \right) + 1,
\end{equation}
where $\xi(a)$ returns the number of arguments of function $a$.

After determining the chromosome length, the next crucial aspect is the representation of individual elements within each chromosome using integers. Four distinct integer ranges are defined to represent the different element types: functions, ADFs, terminals, and input arguments for ADFs. These ranges are specified as $[0, A - 1]$, $[A, B - 1]$, $[B, C - 1]$, and $[C, D - 1]$. The upper bounds $A, B, C,$ and $D$ are calculated as follows:
\begin{align}
A &= \max_{i \in \{1, \dots, K\}} |F_i| ,\label{eq:A_value} \\
B &= A + N_a, \label{eq:B_value} \\
C &= B + \max_{i \in \{1, \dots, K\}} |T_i|,\label{eq:C_value} \\
D &= C + N_g, \label{eq:D_value}
\end{align}
where $|F_i|$ returns the number of elements in set $F_i$, $N_a$ is the number of ADFs in each chromosome, and $N_g$ is the number of input arguments for each ADF.

With this common chromosome representation, the decoding process that transforms a general chromosome into a task-specific chromosome for evaluation is conducted in two distinct phases: 1) Element type identification determines the corresponding type represented by each integer dimension of the chromosome. This is achieved by checking which predefined integer range the integer belongs to. 2) Task-specific mapping scales the integer value according to the maximum number of elements for the identified type within the specific task. The scaling formula is given by:
\begin{equation}\label{eq:scaling_formula}
x'_i = \frac{x_i - L_x}{U_x - L_x} \cdot N_x,
\end{equation}
where $[L_x, U_x - 1]$ is the integer range of the element type identified in the first phase, and $N_x$ is the maximum number of elements for this identified type in the context of the current task. Through this mapping, $x'_i$ is converted to an integer between $0$ and $N_x - 1$, which then serves as an index to map $x'_i$ to a meaningful symbol (function or terminal) specific to the task being evaluated.

For population initialization, we construct a unified encoding space accommodating all target PDE tasks. Subsequently, a population, denoted as $POP$, of size $N$ is randomly initialized. Each candidate solution for a PDE is encoded as a vector, with individual elements represented by integers. To differentiate the semantic meaning of these elements, the integer space is partitioned into the four distinct segments detailed above. All elements are randomly sampled from their respective segments according to their predefined types.

\textbf{b) Population Reproduction.} The reproduction process aims to encourage the discovery and implicit transfer of useful genetic material across tasks, allowing beneficial traits found for one task to improve the search for others. To achieve this, NMIPS employs two crossover and mutation strategies: 1) one-point crossover and uniform mutation, and 2) the differential evolution-based mutation and crossover. The two crossover and mutation strategies are randomly selected based on a probability of $rmp$. 

For one-point crossover and uniform mutation, a randomly selected individual {$\mathbf{z}_{r1}$} is paired with the current parent individual {$\mathbf{z}_{i}$}. A single-point crossover operation is performed between them, producing an intermediate offspring {$\mathbf{y}_{i}$}. This process facilitates the exchange of potentially useful genetic material. Subsequently, a uniform mutation is applied to {$\mathbf{y}_{i}$} to generate the final offspring {$\mathbf{u}_{i}$}, which further enhances population diversity and aids in escaping local optima. The offspring's skill factor $\tau$ is randomly inherited from either {$\mathbf{z}_{i}$} or {$\mathbf{z}_{r1}$} with equal probability.

Alternatively, if the one-point crossover and uniform mutation are not performed, the differential evolution strategy constructs a differential vector using the best individual {$\mathbf{z}_{best}$} from the current population and two other distinct random individuals. This vector is then scaled by a mutation rate $\epsilon$. A gene-level crossover is then performed between the target individual {$\mathbf{z}_{i}$} and this mutation vector. This process generates an offspring that not only inherits partial information from the parent but also incorporates significant new genetic diversity.


Following these reproduction processes, the newly generated offspring population is merged with the parent population to form a temporary population, denoted as $POP_{temp}$, where the skill factor and scalar fitness of each individual are subsequently updated.

\textbf{c) Population Evaluation.} The evaluation process must consider not only the data point fitting error but also the underlying physical consistency dictated by PDEs. Specifically, for a given individual {$\mathbf{z}$} on task $\mathcal{T}_i$, its objective function value, denoted as {$f_i(\mathbf{z})$}, serves as the factorial cost in multifactorial optimization. This {$f_i(\mathbf{z})$} is computed by combining both the data point fitting error and the PDE fitting error for task $\mathcal{T}_i$.

For the data point fitting error, the NMIPS utilizes the root mean square error (RMSE) as an evaluation metric to quantify the candidate solution's performance in fitting the training samples. Given a set of data points $\{(x_k, t_k, u_k)\}_{k=1}^{N_d}$ sampled from a high-precision numerical simulation of a particular PDE, where $(x_k, t_k)$ represents the coordinate point and $u_k$ is the corresponding true solution value at that point. If the generated candidate analytical solution expression is $\hat{u}(x, t)$, then the formula for calculating the data point fitting error on these sampled points is as follows:
\begin{equation}
    \mathcal{L}_{\text{data}} = \sqrt{\frac{1}{N_d} \sum_{k=1}^{N_d} (\hat{u}(x_k, t_k) - u_k)^2},
\end{equation}
where $N_d$ denotes the total number of sampled data points used for training.

To ensure the candidate expression satisfies the physical constraints of the original PDE, we incorporate the PDE fitting error. This comprehensive error term comprises three distinct components: the PDE residual error, the initial condition error, and the boundary condition error. Let the form of the PDE be expressed as:
\begin{equation}
F(x, t, u, \frac{\partial u}{\partial x}, \frac{\partial u}{\partial t}, \dots) = 0,
\end{equation}
where $F(u, \frac{\partial u}{\partial x}, \frac{\partial u}{\partial t}, \dots)$ represents the PDE expression constructed from the unknown function $u$ and its partial derivatives of various orders. After substituting the candidate expression $\hat{u}(x, t)$ into the target PDE, the PDE residual can be calculated on a set of test points $\{(x_j, t_j)\}_{j=1}^{N_{f}}$, defined as:
\begin{equation}
\mathcal{L}_{\text{PDE\_residual}} = \sqrt{\frac{1}{N_{f}} \sum_{j=1}^{N_{f}} |F(x_j, t_j, \hat{u}, \frac{\partial \hat {u}}{\partial x}, \frac{\partial \hat{u}}{\partial t}, \dots)|^2},
\end{equation}
where $N_{f}$ denotes the number of collocation points sampled within the spatiotemporal domain to evaluate the physics residual.

Furthermore, to guarantee the consistency of the candidate expression at the initial temporal moment and spatial boundaries, initial condition error and boundary condition error terms are introduced separately. Taking the initial condition $u(x, 0) = f(x)$ as an illustrative example, its corresponding error can be formulated as:
\begin{equation}
\mathcal{L}_{\text{IC}} = \sqrt{\frac{1}{N_{ic}} \sum_{k=1}^{N_{ic}} (\hat{u}(x_k, 0) - f(x_k))^2},
\end{equation}
where $N_{ic}$ denotes the number of sampling points distributed along the initial boundary.
Similarly, the boundary condition error $\mathcal{L}_{\text{BC}}$ can be constructed, thereby collectively forming the complete PDE fitting error component.

To ensure a fair assessment of the discovered structures, we optimize numerical constants prior to assigning the final fitness. Specifically, constants within each expression are treated as learnable parameters and fine-tuned via a gradient-based optimizer to minimize the combined loss from data and PDE constraints. We employ automatic differentiation to construct computational graphs, enabling the precise computation of high-order derivatives required for both constant tuning and residual evaluation.


\textbf{d) Population Selection.} Based on the factorial costs across all tasks, the following multifactorial metrics are derived to determine which individuals from the temporary population ($POP_{temp}$) advance to the next generation:

\begin{itemize}
    \item Factorial rank ($r_{\mathbf{z}}^j$): For an individual $\mathbf{z}$, its factorial rank $r_{\mathbf{z}}^j$ on task $\mathcal{T}_j$ is its position when the entire population is sorted in ascending order of factorial cost $f_j(\mathbf{z})$ for task $\mathcal{T}_j$. A lower rank indicates better performance.
    
    \item Skill factor ($\tau_{\mathbf{z}}$): The skill factor $\tau_{\mathbf{z}}$ of an individual $\mathbf{z}$ indicates the task on which it performs best (i.e., achieves the lowest factorial rank). It is defined as $\tau_{\mathbf{z}}= \arg\min_j \{ r_{\mathbf{z}}^j \}$.
    
    \item Scalar fitness ($\varphi_{\mathbf{z}}$): The scalar fitness $\varphi_{\mathbf{z}}$ is a task-independent metric used to compare individuals across different tasks. It is defined as $\varphi_{\mathbf{z}} = 1 /[{\min_{j \in \{1, \dots, K\}} r_{\mathbf{z}}^j}]$. This metric consolidates multi-task performance into a single value, where a higher scalar fitness implies better overall performance across its best-performing task. Formally, an individual $\mathbf{z}_a$ is said to multifactorially dominate $\mathbf{z}_b$ if $\varphi_{\mathbf{z}_a} > \varphi_{\mathbf{z}_b}$.
\end{itemize}

To select individuals from the population for survival into the next generation, we adopt a one-to-one selection strategy, commonly employed in the differential evolution algorithm. This mechanism critically depends on a scalar fitness metric derived during the population evaluation phase. 
Specifically, each newly generated offspring {$\mathbf{u}_i$} undergoes a direct comparison against its corresponding parent {$\mathbf{z}_i$}. The update rule for the $i$-th individual in the next generation is defined as:
\begin{equation}
    \mathbf{z}_i = 
    \begin{cases} 
    \mathbf{u}_i, & \text{if } \varphi_{\mathbf{u}_i} > \varphi_{\mathbf{z}_i} ~~\text{or}~~ \mathbf{z}_i \in \mathcal{S}_{\text{redundant}} \\
    \mathbf{z}_i, & \text{otherwise}
    \end{cases}
\end{equation}
where $\mathcal{S}_{\text{redundant}}$ represents the set of redundant individuals (e.g., duplicate solutions or non-viable candidates). If {$\mathbf{u}_i$} exhibits a superior scalar fitness compared to {$\mathbf{z}_i$}, or if {$\mathbf{z}_i$} is identified as a redundant individual, then {$\mathbf{u}_i$} successfully survives and replaces {$\mathbf{z}_i$} in the subsequent generation. Conversely, if neither of these conditions is met, {$\mathbf{z}_i$} is retained in the population. This selective strategy serves a dual purpose: it not only aids in preserving high-quality solutions but also dynamically contributes to enhancing population diversity by promptly replacing less fit or redundant individuals.

\subsection{Affine Transformation for Knowledge Transfer}

To bolster the knowledge sharing across diverse PDE parameter configurations, we propose an affine transformation-based knowledge transfer module. As illustrated in Fig.~\ref{fig:framework} (b), this module operates on a dual-population mechanism, designed to align the statistical distributions of two distinct populations (denoted as $\mathcal{Z}_1$ and $\mathcal{Z}_2$) via learnable scaling and shifting operations. Affine transformations are notably adept at facilitating this alignment while effectively preserving the topological structure of the original features. Consequently, we leverage this approach to convey valuable genetic material and solution structures from the source task's search space to that of the target task, thereby enhancing structural diversity.

Let the two populations from different tasks or evolutionary branches be represented as $\mathcal{Z}_1 = \{\mathbf{z}_1^{(i)}\}_{i=1}^{N}$ and $\mathcal{Z}_2 = \{\mathbf{z}_2^{(j)}\}_{j=1}^{N}$. To capture the global distribution characteristics, we compute the mean vector $\boldsymbol{\mu}_k$ and variance vector $\boldsymbol{\sigma}_k^2$ for each population $k \in \{1, 2\}$. These statistics serve as inputs for the subsequent alignment step.

\begin{algorithm}[t]
    \small
    \caption{Affine Transformation-based Knowledge Transfer}
    \label{alg: KnowledgeTransfer}
    \begin{algorithmic}[1]
        \renewcommand{\algorithmicrequire}{\textbf{Input:}}
        \renewcommand{\algorithmicensure}{\textbf{Output:}}
        
        \Require Current population $\mathcal{Z}$, Population size $N$.
        \Ensure Transferred population $\mathcal{Z}_{new}$.
        \State Divide $\mathcal{Z}$ into task-specific groups $G_1$ and $G_2$ according to skill factor $\tau$;
        \State Sort $G_1, G_2$ descendingly by scalar fitness $\varphi$;
        \State $\mathcal{Z}_1 \leftarrow$ Top $N/2$ individuals from $G_1$;
        \State $\mathcal{Z}_2 \leftarrow$ Top $N/2$ individuals from $G_2$;
        
        \For{$k = 1$ to $2$}
            \State Calculate mean $\boldsymbol{\mu}_k$ and variance $\boldsymbol{\sigma}_k^2$ of $\mathcal{Z}_k$;
            \State Obtain affine parameters $\gamma_k,~\beta_k$ by Eq.~(\ref{mlp});
        \EndFor
        
        \State $\mathcal{Z}'_1 \leftarrow$ Apply affine transform on $\mathcal{Z}_1$ using $\boldsymbol{\mu}_1, \boldsymbol{\sigma}_1, \gamma_1, \beta_1$;
        \State $\mathcal{Z}'_2 \leftarrow$ Apply affine transform on $\mathcal{Z}_2$ using $\boldsymbol{\mu}_2, \boldsymbol{\sigma}_2, \gamma_2, \beta_2$;

        \State Repair genes in $\mathcal{Z}'_1 \cup \mathcal{Z}'_2$ via modulo arithmetic to valid integer ranges;
        
        \State $\mathcal{Z}_{temp} = \mathcal{Z}_1 \cup \mathcal{Z}_2 \cup \mathcal{Z}'_1 \cup \mathcal{Z}'_2$;
        \State Re-evaluate factorial costs and update $\varphi$ for all $\mathbf{z}\in \mathcal{Z}_{temp}$;
        \State Sort $\mathcal{Z}_{temp}$ descendingly by scalar fitness $\varphi$;
        \State $\mathcal{Z}_{new} \leftarrow$ Top $N$ individuals from $\mathcal{Z}_{temp}$;
        
        \State \Return $\mathcal{Z}_{new}$;
    \end{algorithmic}
\end{algorithm}


To map the statistics of $\mathcal{Z}_1$ towards $\mathcal{Z}_2$ and vice versa, we employ two parallel multi-layer perceptrons ($\text{MLP}_1$ and $\text{MLP}_2$). These networks output two learnable parameters for affine transformation, a scaling factor $\gamma$ and an offset factor $\beta$, which modulate the mapping between $\mathcal{Z}_1$ and $\mathcal{Z}_2$. Such transformations have been proven to effectively promote the resolution of PDEs within PINNs \cite{mandl2023affine}. By feeding the extracted moments into the MLPs, the parameters are optimized in a data-driven manner to minimize distribution discrepancy.
Formally, the mapping process for the $k$-th population is defined as:
\begin{equation}
[\gamma_k, \beta_k] = \text{MLP}_k(\boldsymbol{\mu}_k, \boldsymbol{\sigma}_k^2; \boldsymbol{\theta}_k), \label{mlp}
\end{equation}
where $\boldsymbol{\theta}_k$ represents the trainable weights of the MLP. 

With the learned parameters $\gamma_k$ and $\beta_k$, we apply an affine transformation to the normalized individuals. For an individual $\mathbf{z}_k \in \mathcal{Z}_k$, the transformed individual $\mathbf{z}'_k$ is given by:
\begin{equation}
\mathbf{z}'_k = (1 + \gamma_k) \odot \frac{\mathbf{z}_k - \boldsymbol{\mu}_k}{\sqrt{\boldsymbol{\sigma}_k^2 + \varepsilon}} + \beta_k,
\end{equation}
where $\odot$ denotes element-wise multiplication, and $\varepsilon$ is a small constant for numerical stability. This operation effectively projects the individuals into a new search space that fuses the structural properties of the original population with the statistical characteristics of the counterpart. 

To ensure the transformed distribution effectively aligns with the target distribution, we employ MSE as the optimization objective. We minimize the element-wise difference between the transformed source population and the target population, which acts as a proxy for minimizing the Wasserstein distance between the two distributions. The loss function $\mathcal{L}_{align}$ is defined as:
\begin{equation}
    \mathcal{L}_{align} = \frac{1}{N} \sum_{i=1}^{N} \| \mathbf{z}'_i - \mathbf{z}_{target}^{(i)} \|^2,
\end{equation}
where $\mathbf{z}_{target}^{(i)}$ represents the $i$-th individual in the target population. 
Finally, to leverage the diversity gained from this bidirectional transfer and drive a renewed selection process, the transformed populations are merged with the original pools. The aggregated population $\mathcal{Z}_{temp}$ is formed by concatenating the transformed sets, denoted as:
\begin{equation}
    \mathcal{Z}_{temp} = \mathcal{Z}_1 \cup \mathcal{Z}_2 \cup \mathcal{Z}'_1 \cup \mathcal{Z}'_2. 
\end{equation}
A rigorous selection process is then applied to $\mathcal{Z}_{temp}$ to identify superior individuals for the next evolutionary cycle. This knowledge transfer enhances population diversity and significantly improves the evolutionary efficiency of the target task. The complete procedure is detailed in Algorithm \ref{alg: KnowledgeTransfer}.

\section{EXPERIMENTS}
\label{experiment}

\subsection{Experiments setup}

\subsubsection{Datasets}
To evaluate the effectiveness of NMIPS in solving parameterized PDEs, we constructed a benchmark dataset consisting of six representative PDEs: the 1D Advection equation, 1D Burgers' equation, 1D Advection-Diffusion equation, 2D Advection equation, 2D Navier-Stokes equation, and 3D Advection equation.
For each PDE, we systematically varied scalar and functional parameters to generate multiple tasks. For each PDE, while keeping the equation structure unchanged, four different sets of functional parameters were sampled to form the corresponding PDE Family. For each task, 1100 points were randomly sampled from the computational domain to construct the training and evaluation datasets. For PDEs with known analytical solutions, data were generated using closed-form expressions. For PDEs lacking general analytical solutions, high-precision numerical methods were employed to compute reference solutions.

\subsubsection{Baselines}
To validate the performance of NMIPS, we compared it with five state-of-the-art baselines: PhySO~\cite{tenachi2023deep}, DSR~\cite{petersen2021deep}, SP-GPSR~\cite{cao2023genetic}, GNOT~\cite{hao2023gnot}, and Geo-FNO~\cite{li2023fourier}. These baselines span various approaches, including symbolic regression, genetic programming, and neural operators. 
    
    
    
    

For more detailed experimental settings, please refer to the supplemental material (see Section II). This includes detailed descriptions of each equation in the dataset, specific configurations of the baseline methods, as well as the hyper-parameter settings (e.g., population size, mutation rate) and the symbol library used for NMIPS.

\subsection{Results and Discussion}

\subsubsection{Results on the 1D Advection Equation}

To evaluate the predictive accuracy of NMIPS and compare it with existing approaches, we conducted experiments on the 1D advection equation with varying advection coefficients, $\beta$. The mean squared error (MSE) of the predicted physical fields is presented in Table \ref{base:adv1d}. The table includes results from NMIPS under different combinations (Num. 0, 1, 2, 3) of $\beta$ of two PDEs and five other baseline methods. The results reveal several key insights. First, NMIPS consistently demonstrates superior performance across a range of advection coefficients. Specifically, NMIPS achieves an average MSE of 5.47E-01, which is lower than all other baselines. This indicates that NMIPS can more accurately model the physical field evolution of the 1D advection equation compared to existing state-of-the-art methods. Second, while the performance of all models generally fluctuates with changes in $\beta$, NMIPS exhibits a more stable and consistently low error rate. The overall average MSE confirms that NMIPS not only achieves competitive performance but also demonstrates a superior level of generalization across different physical parameter settings. 

To provide a visual assessment of model fidelity, the spatial-temporal error distributions for the specific case of $\beta=1.000$ are presented in supplemental material (see Fig. S-7). These visualizations corroborate the quantitative results, confirming that NMIPS maintains significantly higher fidelity to the exact dynamics compared to baseline methods, which exhibit more pronounced deviations.


\begin{table}[t]
  \centering
  \caption{Discovered expressions for the 1D Burgers' equation by different symbolic regression methods}
  \begin{tabular}{cc}
    \toprule
    \textbf{Method} & \textbf{Expression} \\
    \midrule
    NMIPS & $u = 0.014(t - x) + 0.024$ \\
    PhySO & $u = \sin(16809.645e^{-105.420t})$ \\
    DSR & $u = \sin(e^{ t \cdot (1.0 - e^{e^{e^{x}}})})$ \\
    SP-GPSR & $u = 3.496e^{-5.511e^{-e^{-10.0t}}}$ \\
    \bottomrule
  \end{tabular}
  \label{tab:burgers_sol}
\end{table}

\begin{table*}[t]
    \centering
    \setlength{\tabcolsep}{5pt}
    \caption{MSE of different methods on 1D Advection equation}
    \begin{tabular}{c w{c}{0.8cm} *{9}{w{c}{1cm}}}
        \toprule
        \multirow{2}*{\textit{\textbf{Num}}} & \multirow{2}*{\textbf{$\beta$}} & \multicolumn{4}{c}{\textbf{NMIPS}} & \multirow{2}*{\textbf{PhySO}} & \multirow{2}*{\textbf{DSR}} & \multirow{2}*{\textbf{SP-GPSR}} & \multirow{2}*{\textbf{GNOT}} & \multirow{2}*{\textbf{Geo-FNO}} \\
         & & 0 & 1 & 2 & 3 & & & & & \\
        \midrule
        0 & 0.100  & / & 5.64E-01 & 5.87E-01 & 5.75E-01 & 6.24E-01 & 5.93E-01 & 6.08E-01 & 8.18E-01 & 9.52E-01 \\
        1 & 0.400  & 4.34E-01 & / & 4.33E-01 & 4.33E-01 & 4.64E-01 & 4.51E-01 & 4.44E-01 & 4.48E-01 & 5.39E-01 \\
        2 & 0.700  & 1.46E-01 & 1.48E-01 & / & 1.60E-01 & 1.59E-01 & 1.59E-01 & 1.51E-01 & 5.25E-01 & 2.14E-01 \\
        3 & 1.000  & 1.02E+00 & 1.04E+00 & 1.02E+00 & / & 1.14E+00 & 1.03E+00 & 1.11E+00 & 6.55E-01 & 1.17E+00 \\
        \midrule
        \multicolumn{2}{c}{\textbf{Avg}} & \multicolumn{4}{c}{\textbf{5.47E-01}}  & 5.97E-01 & 5.58E-01 & 5.77E-01 & 6.11E-01 & 7.19E-01 \\
        \bottomrule
    \end{tabular}
    \label{base:adv1d}
\end{table*}

\begin{table*}[t]
    \centering
    \setlength{\tabcolsep}{5pt}
    \caption{MSE of different methods on 1D Burgers' equation}
    \begin{tabular}{c w{c}{0.8cm} *{9}{w{c}{1cm}}}
        \toprule
        \multirow{2}*{\textit{\textbf{Num}}} & \multirow{2}*{\textbf{$\nu$}} & \multicolumn{4}{c}{\textbf{NMIPS}} & \multirow{2}*{\textbf{PhySO}} & \multirow{2}*{\textbf{DSR}} & \multirow{2}*{\textbf{SP-GPSR}} & \multirow{2}*{\textbf{GNOT}} & \multirow{2}*{\textbf{Geo-FNO}} \\
         & & 0 & 1 & 2 & 3 & & & & & \\
        \midrule
        0 & 0.001  & / & 3.98E-02 & 3.97E-02 & 4.04E-02 & 7.52E-02 & 4.77E-02 & 4.51E-02 & 2.00E-01 & 5.72E-02 \\
        1 & 0.004  & 6.39E-03 & / & 7.01E-03 & 6.50E-03 & 5.59E-03 & 6.31E-03 & 7.19E-03 & 8.37E-02 & 1.22E-02 \\
        2 & 0.007  & 7.14E-03 & 6.20E-03 & / & 6.37E-03 & 5.40E-03 & 5.92E-03 & 5.68E-03 & 5.23E-02 & 1.13E-02 \\
        3 & 0.010  & 1.02E-02 & 6.94E-03 & 7.00E-03 & / & 8.98E-03 & 7.90E-03 & 7.31E-03 & 7.92E-02 & 1.08E-02 \\
        \midrule
        \multicolumn{2}{c}{\textbf{Avg}} & \multicolumn{4}{c}{\textbf{1.53E-02}}  & 2.38E-02 & 1.69E-02 & 1.63E-02 & 1.04E-01 & 2.29E-02 \\
        \bottomrule
    \end{tabular}
    \label{base:burgers1d}
\end{table*}

\begin{figure*}[t]
    \centerline{\includegraphics[scale=0.38]{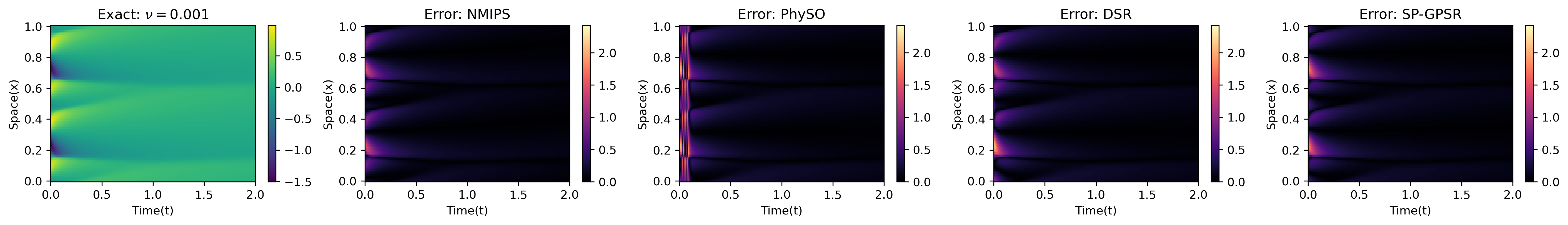}}
    \caption{A spatial-temporal error analysis of different methods for the 1D Burgers' equation.}
    \label{fig:burgers_error}
\end{figure*}


\subsubsection{Results on the 1D Burgers' Equation}

To evaluate the predictive accuracy of NMIPS on the 1D Burgers' equation, we conducted experiments with varying viscosity coefficients, $\nu$. As shown in Table \ref{base:burgers1d}, NMIPS achieves the best overall performance with an average MSE of 1.53E-02. This result outperforms the closest competitors, SP-GPSR (1.63E-02) and DSR (1.69E-02), while maintaining a significant lead over other comparative methods. 

A key observation is our method's stable performance across different values of $\nu$. For instance, at $\nu=0.001$, our method achieves a minimum MSE of 3.97E-02, outperforming all competitors. Although some baselines (e.g., PhySO at $\nu=0.007$) may achieve a lower MSE at a specific point, our method maintains consistent accuracy across all tested $\nu$ values, as evidenced by our overall average.
Furthermore, qualitative analysis further corroborates these quantitative findings. The spatial-temporal error maps in Fig.~\ref{fig:burgers_error} ($\nu=0.001$) show that NMIPS yields significantly lower pointwise error—visualized as darker regions—particularly in high-gradient shock wave zones. The symbolic expressions in Table~\ref{tab:burgers_sol} offer insight into this performance gap. While NMIPS identifies a concise and physically interpretable linear relation ($u \propto t-x$), the baseline methods converge on overly complex expressions containing nested exponentials. This indicates that NMIPS effectively captures the true underlying dynamics, avoiding the overfitting issues commonly observed in other symbolic regression approaches.

\subsubsection{Results on the 1D Advection-Diffusion Equation}

To evaluate the capabilities of NMIPS on the 1D Advection-Diffusion equation, we conducted experiments across various combinations of the advection coefficient ($\alpha$) and diffusion coefficient ($\beta$). As detailed in the supplemental material (see Table S-I), NMIPS achieves the lowest average MSE of 1.19E-01. This performance surpasses all baseline methods, including the closest competitors, SP-GPSR\ignore{(1.25E-01)} and DSR\ignore{(1.26E-01)}, demonstrating superior overall accuracy.
The specific challenge of this equation lies in regimes where advection dominates diffusion. In such scenarios, rapid transport creates steep gradients that are not smoothed by diffusion, leading to shock-like features that are notoriously difficult to model. For the challenging setting of $\alpha=0.700$ and $\beta=0.001$, NMIPS achieves a competitive MSE of 1.47E-01, significantly outperforming PhySO (1.71E-01) and Geo-FNO (1.56E-01).

The error heatmaps in the supplemental material (see Fig. S-8) corroborate the quantitative findings, further underscoring NMIPS's strong generalization capabilities in effectively balancing the dynamics of advection and diffusion.


\subsubsection{Results on the 2D Advection Equation}

To evaluate the predictive accuracy of NMIPS on the 2D advection equation, we conducted experiments with various combinations of advection coefficients, $\beta_x$ and $\beta_y$. As presented in Table \ref{base:adv2d}, NMIPS achieves the best overall performance with a low average MSE of 2.58E-01. This result surpasses closely competing symbolic regression methods such as PhySO and SP-GPSR, while significantly outperforming deep learning baselines like GNOT and Geo-FNO.

The results further highlight the robustness of NMIPS across different flow regimes. For the computationally challenging case characterized by high advection in both dimensions ($\beta_x=0.700,~\beta_y=0.969$), our method achieves a minimum MSE of 2.41E-01, demonstrating its ability to model complex multidimensional dynamics effectively. In contrast, baselines such as GNOT exhibit extreme performance fluctuations; while GNOT achieves low error in specific low-advection settings (e.g., Num 0), its error spikes drastically in other cases, resulting in poor overall generalization. Conversely, NMIPS maintains consistent stability across all parameter combinations, validating its superior adaptability. Additionally, the error heatmap for a specific case is presented in the supplemental material (see Fig. S-9).

\begin{table*}[t]
    \centering
    \setlength{\tabcolsep}{5pt}
    \caption{MSE of different methods on 2D Advection equation}
    \begin{tabular}{cccccccccccc}
        \toprule
        \multirow{2}*{\textit{\textbf{Num}}} & \multirow{2}*{\textbf{$\beta_x$}} & \multirow{2}*{\textbf{$\beta_y$}} & \multicolumn{4}{c}{\textbf{NMIPS}} & \multirow{2}*{\textbf{PhySO}} & \multirow{2}*{\textbf{DSR}} & \multirow{2}*{\textbf{SP-GPSR}} & \multirow{2}*{\textbf{GNOT}} & \multirow{2}*{\textbf{Geo-FNO}} \\
        & & & 0 & 1 & 2 & 3 & & & & & \\
        \midrule
        0 & 0.100  & 0.842  & / & 2.86E-01 & 2.55E-01 & 2.92E-01 & 2.81E-01 & 2.63E-01 & 2.81E-01 & 1.38E-01 & 4.36E-01 \\
        1 & 0.400  & 0.349  & 2.55E-01 & / & 2.60E-01 & 2.53E-01 & 2.56E-01 & 2.88E-01 & 2.51E-01 & 5.17E-01 & 4.04E-01 \\
        2 & 0.700  & 0.969  & 2.50E-01 & 2.41E-01 & / & 2.45E-01 & 2.53E-01 & 2.72E-01 & 2.65E-01 & 5.19E-01 & 3.92E-01 \\
        3 & 1.000  & 0.186  & 2.50E-01 & 2.56E-01 & 2.57E-01 & / & 2.67E-01 & 2.60E-01 & 2.62E-01 & 5.03E-01 & 4.19E-01 \\
        \midrule
        \multicolumn{3}{c}{\textbf{Avg}} & \multicolumn{4}{c}{\textbf{2.58E-01}} & 2.64E-01 & 2.71E-01 & 2.64E-01 & 4.19E-01 & 4.13E-01 \\
        \bottomrule
    \end{tabular}
    \label{base:adv2d}
\end{table*}

\begin{table*}[htbp]
    \centering
    \setlength{\tabcolsep}{5pt}
    \caption{MSE of different methods on 3D Advection equation}
    \begin{tabular}{ccccccccccccc}
        \toprule
        \multirow{2}*{\textit{\textbf{Num}}} & \multirow{2}*{\textbf{$\beta_x$}} & \multirow{2}*{\textbf{$\beta_y$}} & \multirow{2}*{\textbf{$\beta_z$}} & \multicolumn{4}{c}{\textbf{NMIPS}} & \multirow{2}*{\textbf{PhySO}} & \multirow{2}*{\textbf{DSR}} & \multirow{2}*{\textbf{SP-GPSR}} & \multirow{2}*{\textbf{GNOT}} & \multirow{2}*{\textbf{Geo-FNO}} \\
        & & & & 0 & 1 & 2 & 3 & & & & & \\
        \midrule
        0 & 0.100  & 0.983  & 0.548  & / & 1.23E-01 & 1.24E-01 & 1.30E-01 & 1.32E-01 & 1.41E-01 & 1.33E-01 & 4.07E-01 & 2.11E-01 \\
        1 & 0.400  & 0.579  & 0.573  & 1.27E-01 & / & 1.27E-01 & 1.46E-01 & 1.36E-01 & 1.39E-01 & 1.31E-01 & 3.57E-01 & 2.02E-01 \\
        2 & 0.700  & 0.818  & 0.951  & 1.23E-01 & 1.27E-01 & / & 1.26E-01 & 1.46E-01 & 1.40E-01 & 1.41E-01 & 3.63E-01 & 1.93E-01 \\
        3 & 1.000  & 0.697  & 0.204  & 1.38E-01 & 1.33E-01 & 1.34E-01 & / & 1.40E-01 & 1.40E-01 & 1.34E-01 & 3.59E-01 & 1.84E-01 \\
        \midrule
        \multicolumn{4}{c}{\textbf{Avg}} & \multicolumn{4}{c}{\textbf{1.30E-01}} & 1.39E-01 & 1.40E-01 & 1.35E-01 & 3.72E-01 & 1.97E-01 \\
        \bottomrule
    \end{tabular}
    \label{base:adv3d}
\end{table*}

\subsubsection{Results on the 2D Navier-Stokes Equation}


To evaluate the predictive accuracy of NMIPS on the 2D Navier-Stokes equation, we conducted experiments across a range of viscosity coefficients ($\nu$). As presented in the supplemental material (see Table S-II), NMIPS demonstrates leading performance. With an average MSE of 5.25E-02, NMIPS outperforms all other baselines, including PhySO, Geo-FNO, and GNOT, and maintains a competitive edge over SP-GPSR and DSR. This confirms our model's effectiveness in accurately capturing the complex dynamics of viscous fluid flow.

The robustness of NMIPS is particularly evident in specific flow regimes. For instance, at $\nu=0.020$, it achieves a minimum MSE of 4.36E-02, establishing a clear performance margin over PhySO (1.14E-01) and GNOT (1.07E-01). For a deeper investigation into the source of this accuracy, we provide the discovered symbolic expressions (see Table S-III) and their detailed analysis in the supplemental material. Additionally, the error heatmap for a representative case is visualized in the supplemental material (see Fig. S-10), further corroborating the high fidelity of our approach.

\subsubsection{Results on the 3D Advection Equation}

To evaluate the predictive accuracy of NMIPS on the 3D advection equation, we conducted experiments with varying advection coefficients ($\beta_x$, $\beta_y$, and $\beta_z$). As shown in Table \ref{base:adv3d}, NMIPS consistently achieves superior performance. With an average MSE of 1.30E-01, it outperforms all baselines. The robustness of NMIPS is particularly evident in high-advection regimes. For the challenging case characterized by high coefficients across all three dimensions ($\beta_x=0.700,~\beta_y=0.818,~\beta_z=0.951$), NMIPS maintains a low MSE of 1.23E-01. This performance is notably better than that of PhySO (1.46E-01) and DSR (1.40E-01), highlighting the model's strong generalization capabilities under complex transport conditions.

This quantitative superiority is further supported by the symbolic analysis provided in the supplemental material (see Table S-IV). NMIPS identifies a physically plausible form that correctly captures the advective relationship, whereas baseline methods fail to identify the correct symbolic form.



\subsubsection{Results on Computational Costs}

To evaluate computational efficiency, we compared NMIPS with \ignore{other leading} symbolic regression baselines across six benchmark PDE tasks. As shown in the supplemental material (see Fig. S-11), the vertical axis is plotted on a logarithmic scale to clearly highlight the orders-of-magnitude differences in runtime among the methods.

The figure clearly demonstrates that NMIPS holds a significant computational advantage, finishing as the fastest approach in the majority of tasks, including the 1D Burgers', 1D Advection-Diffusion, 2D Navier-Stokes, and 3D Advection equations. 
The difference is particularly stark for the 1D tasks. For the 1D Burgers' and 1D Advection-Diffusion equations, NMIPS completes in approximately 250-300 seconds. In contrast, competitors like PhySO, DSR, and SP-GPSR require runtimes in the thousands of seconds (ranging from 2000s to 4000s), representing a full order-of-magnitude speedup for our approach. 
While PhySO and SP-GPSR proved faster on the 1D Advection and 2D Advection tasks, respectively, NMIPS remained highly competitive. More importantly, NMIPS consistently avoids the extreme computational costs that plague other symbolic methods in complex scenarios.

Overall, these results validate the computational efficiency of NMIPS. When combined with the high accuracy and interpretability demonstrated in previous sections, NMIPS establishes its practicality and ability to deliver a more balanced trade-off between accuracy, speed, and interpretability than competing symbolic regression approaches.

\begin{table*}[t]
    \centering
    \setlength{\tabcolsep}{5pt}
    \caption{MSE on different equations with and without affine transformation}
    \begin{tabular}{ccccccccccccc}
        \toprule
        \multirow{2}{*}{\textit{\textbf{Num}}} & \multicolumn{2}{c}{\textbf{1D Advection}} & \multicolumn{2}{c}{\textbf{1D Burgers'}} & \multicolumn{2}{c}{\textbf{1D Advection-Diffusion}} & \multicolumn{2}{c}{\textbf{2D Advection}} & \multicolumn{2}{c}{\textbf{2D Navier-Stokes}} & \multicolumn{2}{c}{\textbf{3D Advection}} \\
        & \textbf{w/ AT} & \textbf{w/o AT} & \textbf{w/ AT} & \textbf{w/o AT} & \textbf{w/ AT} & \textbf{w/o AT} & \textbf{w/ AT} & \textbf{w/o AT} & \textbf{w/ AT} & \textbf{w/o AT} & \textbf{w/ AT} & \textbf{w/o AT} \\
        \midrule
        0 & 5.75E-01 & 7.31E-01 & 3.99E-02 & 4.30E-02 & 2.66E-01 & 2.77E-01 & 2.78E-01 & 3.35E-01 & 1.23E-01 & 1.30E-01 & 1.26E-01 & 1.32E-01 \\
    	1 & 4.33E-01 & 4.58E-01 & 6.63E-03 & 6.82E-03 & 1.08E-02 & 1.08E-02 & 2.56E-01 & 3.40E-01 & 4.62E-02 & 5.86E-02 & 1.33E-01 & 1.30E-01 \\
    	2 & 1.51E-01 & 1.47E-01 & 6.57E-03 & 6.08E-03 & 1.48E-01 & 1.49E-01 & 2.45E-01 & 2.81E-01 & 2.48E-02 & 3.59E-02 & 1.26E-01 & 1.32E-01 \\
    	3 & 1.03E+00 & 1.09E+00 & 8.06E-03 & 6.98E-03 & 5.13E-02 & 5.12E-02 & 2.54E-01 & 2.96E-01 & 1.64E-02 & 1.75E-02 & 1.35E-01 & 1.92E-01 \\
    	\midrule
    	\textbf{Avg} & \textbf{5.47E-01} & 6.06E-01 & \textbf{1.53E-02} & 1.57E-02 & \textbf{1.19E-01} & 1.22E-01 & \textbf{2.58E-01} & 3.13E-01 & \textbf{5.25E-02} & 6.06E-02 & \textbf{1.30E-01} & 1.47E-01 \\
        \bottomrule
    \end{tabular}
    \label{ablation:at}
\end{table*}

\subsection{Ablation Studies}


To assess the contribution of the knowledge transfer module, we performed an ablation study by removing the affine transformation (w/o AT), a key component for transferring knowledge between different PDEs. The results in Table~\ref{ablation:at} provide a clear comparison of our full model against the ablated version across six distinct physical equations.


The results consistently demonstrate the critical role of the affine transformation module. On average, our full model (w/ AT) consistently outperforms the ablated version (w/o AT) across all six equations. While performance in a few specific, individual parameter settings can fluctuate (e.g., Num 2 for 1D Burgers'), the average MSE metrics confirm the module's consistent and positive contribution.
Quantitatively, the affine transformation provides its most significant performance gain on the 2D Advection equation, reducing the average MSE from 3.13E-01 to 2.58E-01, a substantial 17.6\% improvement. Similarly, notable gains are seen on the 2D Navier-Stokes (a 13.4\% reduction from 6.06E-02 to 5.25E-02) and 3D Advection (an 11.6\% reduction from 1.47E-01 to 1.30E-01) tasks. This strongly suggests that the module's ability to align latent features becomes increasingly crucial as the complexity of the physical system increases.
Even on tasks where the average improvement is more modest, such as the 1D Burgers' (a 2.5\% reduction from 1.57E-02 to 1.53E-02) and 1D Advection-Diffusion (a 2.5\% reduction from 1.22E-01 to 1.19E-01), the module still enhances overall performance. These findings strongly confirm that the affine transformation module successfully facilitates knowledge transfer, enhancing the model's predictive accuracy and generalization. 

\subsection{Sensitivity Studies}

\begin{figure}
    \centering
    \includegraphics[scale=0.5]{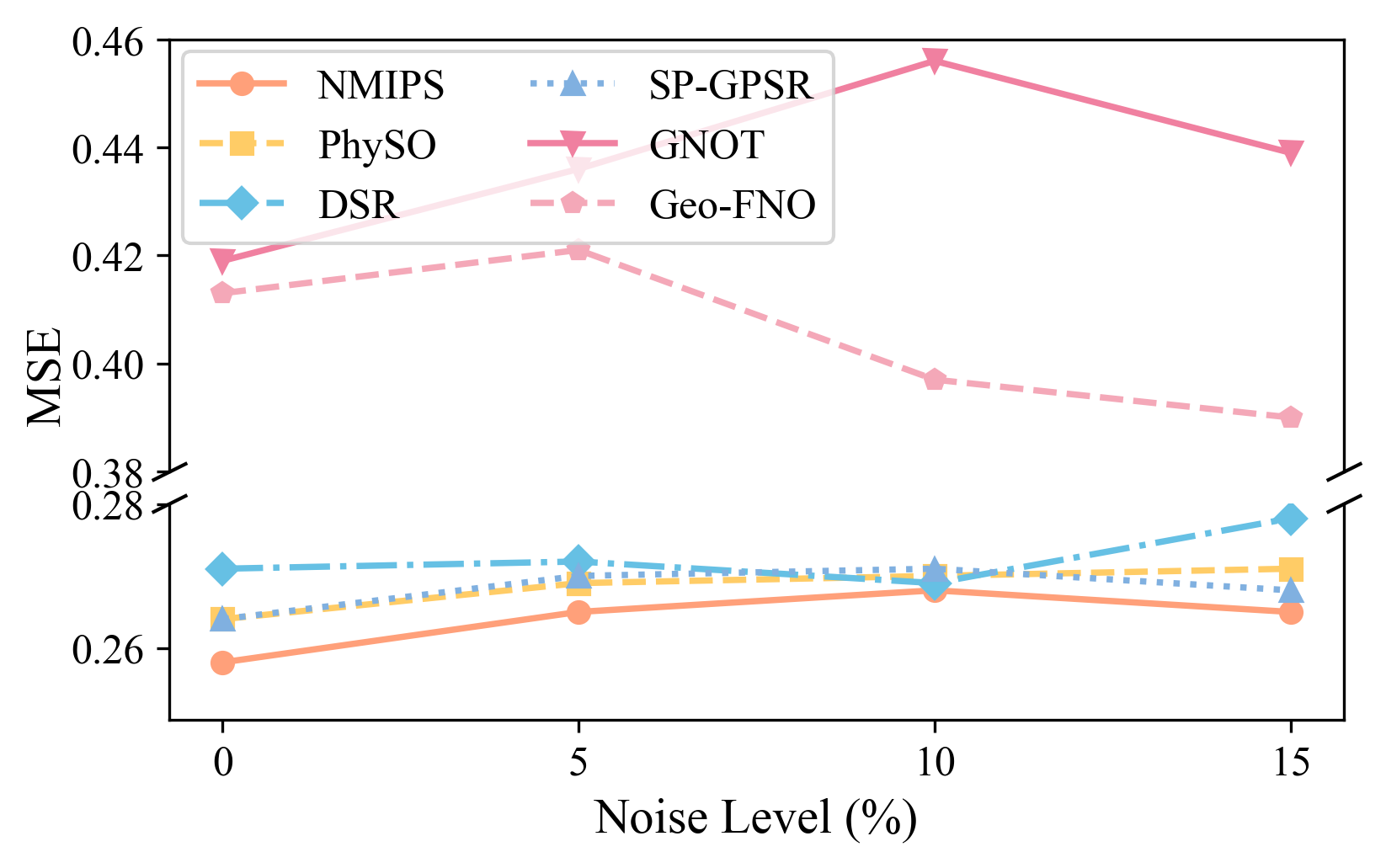}
    \caption{Noise robustness comparison of different methods.}
    \label{fig:noise}
\end{figure}

This section analyzes the performance of the proposed method in predicting physical fields under various noise levels. We add Gaussian noise to both the fields and the interface position data at each sampling point. The noise $\epsilon$ is drawn from a normal distribution $\mathcal{N}(0,\sigma^2)$, where the standard deviation $\sigma$ is set to 5\%, 10\%, and 15\% of the original field magnitude, yielding noisy data $\tilde{u} = u + \epsilon$.

The results, presented in Fig.~\ref{fig:noise}, compare the MSE of our proposed algorithm with five other baselines. The results clearly demonstrate that our method exhibits superior noise robustness. As shown in the graph, our method (solid orange line) consistently achieves the lowest MSE across all tested noise levels, from 0\% to 15\%. Its performance remains exceptionally stable, with the error curve staying almost flat, showing only negligible degradation as data corruption increases. This highlights our model's strong ability to filter out noisy signals and capture the underlying physical principles. In contrast, while other top-performing symbolic regression baselines like PhySO and SP-GPSR also show a high degree of stability, their MSEs are consistently higher than ours at every noise level. DSR, another baseline, shows a more noticeable upward trend in error, indicating a greater sensitivity to increasing noise.

Furthermore, the neural operator models, GNOT and Geo-FNO, appear significantly less robust, exhibiting MSE values on a much higher scale. These models suffer from both lower accuracy and higher volatility. Notably, GNOT's error peaks dramatically at the 10\% noise level. These experimental results validate the remarkable resilience of our method. The ability to maintain consistent accuracy despite data corruption is a critical advantage for real-world applications where data is inherently prone to noise.

\section{CONCLUSION}
\label{conclusion}

Our work introduced a multitasking symbolic regression framework to solve families of PDEs, not only offering speedup but also outputting analytical solutions for interpretability. In particular, the method leverages the shared mathematical structure within a family of PDEs to find an interpretable, analytical solution for all instances simultaneously. 
A novel affine transfer strategy is embedded in the framework to enable effective and rapid transfer of learned features between related PDEs, dramatically accelerating the discovery of analytical solutions compared to learning each instance from scratch. 
Our extensive experimental results confirm the superior performance of our approach. We have demonstrated not only significant improvements in efficiency but also enhanced accuracy and generalization capabilities, particularly in challenging scenarios with limited or noisy data. By delivering a framework that is both computationally efficient and highly interpretable, our work represents a major step toward enabling the widespread discovery of analytical solutions for complex scientific and engineering problems.

\bibliography{mybibtex}

\clearpage

\end{document}


\IEEEiedlistdecl

\title{Supplemental Material: Neuro-Symbolic Multitasking: A Unified Framework for Discovering Generalizable Solutions to PDE Families}

 \author{
    Yipeng~Huang,~\IEEEmembership{}
    Dejun~Xu,~\IEEEmembership{}
    Zexin~Lin,~\IEEEmembership{}
    Zhenzhong~Wang,~\IEEEmembership{Member,~IEEE,}\\
    Min Jiang,~\IEEEmembership{Senior~Member,~IEEE}
    
	\IEEEcompsocitemizethanks{\IEEEcompsocthanksitem
Yipeng Huang, Dejun Xu, Zexin Lin, Min Jiang and Zhenzhong Wang are with the Department of Artificial Intelligence, Key Laboratory of Digital Protection and Intelligent Processing of Intangible Cultural Heritage of Fujian and Taiwan, Ministry of Culture and Tourism, School of Informatics, Xiamen University, and Key Laboratory of Multimedia Trusted Perception and Efficient Computing, Ministry of Education of China, Xiamen University, Xiamen 361005, Fujian, China  \textit{(Corresponding authors: Zhenzhong Wang and Min Jiang, e-mail: zhenzhongwang@xmu.edu.cn and minjiang@xmu.edu.cn)}.

	}	
}
 
    

\bibliographystyle{IEEEtran}
\maketitle

\IEEEdisplaynontitleabstractindextext

\IEEEpeerreviewmaketitle

This supplemental material includes the mathematical background and problem formulation, detailed experimental setup, additional experimental results, scientific insights, and discussion.

\section{mathematical background and problem formulation} \label{pde_bk}

Partial Differential Equations (PDEs) are mathematical equations that involve an unknown function of multiple independent variables and their partial derivatives with respect to those variables. They are fundamental to describing a vast array of physical phenomena across science and engineering, from the propagation of waves and the flow of heat to the dynamics of fluids and the behavior of quantum particles. Unlike ordinary differential equations (ODEs) \cite{delic2025numerical}, which involve functions of a single independent variable, PDEs capture complex interactions and variations across space and time.

Formally, a partial differential equation for an unknown function $u(x_1, x_2, \dots, x_n)$ can be expressed in a general form as:
\begin{equation}
F\left(x_1, \dots, x_n, u, \frac{\partial u}{\partial x_1}, \dots, \frac{\partial u}{\partial x_n}, \frac{\partial^2 u}{\partial x_1^2}, \dots, \frac{\partial^k u}{\partial x_n^k}\right) = 0.
\end{equation}
Here, $u$ represents the unknown function we aim to solve for, depending on multiple independent variables. The terms $x_1, x_2, \dots, x_n$ are the independent variables, often representing spatial coordinates (e.g., $x, y, z$) or time ($t$). Partial derivatives like $\frac{\partial u}{\partial x_i}$ denote the first-order partial derivative of $u$ with respect to $x_i$, and $\frac{\partial^k u}{\partial x_i^k}$ denotes the $k$-th order partial derivative of $u$ with respect to $x_i$; higher-order partial derivatives involving combinations of variables (e.g., $\frac{\partial^2 u}{\partial x_i \partial x_j}$) may also be present. Finally, $F$ is a given function (or operator) that defines the specific relationships between the independent variables, the unknown function, and its partial derivatives, thereby specifying the particular PDE. PDEs are classified based on various properties, including linearity, order (the highest order of derivative present), and type (e.g., elliptic, parabolic, hyperbolic), which often correlate with the physical phenomena they describe.

A PDE family refers to a collection of Partial Differential Equations that share an identical underlying mathematical structure but differ in the specific values of one or more parameters embedded within the equation. This concept is crucial in many practical scenarios where a physical system's fundamental behavior remains constant, but its specific properties or environmental conditions vary.

Formally, a PDE family can be defined as a set of PDEs described by a general functional form $F$ that explicitly includes one or more parameters. Let $\mathbf{p} = (p_1, p_2, \dots, p_m)$ be a vector of $m$ parameters. Then, a PDE belonging to such a family can be written as:
\begin{equation}
G(x_1, \dots, x_n, u, \nabla u, \nabla^2 u, \dots, \nabla^k u; \mathbf{p}) = 0.
\end{equation}
In this definition, $G$ represents the fixed mathematical structure or functional form of the PDE that is common across the entire family, outlining the relationships between the independent variables, the unknown function, and its derivatives. The term $\mathbf{p} = (p_1, p_2, \dots, p_m)$ is the parameter vector, where each unique combination of values for $p_1, \dots, p_m$ specifies a distinct instance of the PDE within the family. These parameters frequently embody physical constants (e.g., diffusion coefficients, reaction rates, material properties like viscosity or thermal conductivity) or external environmental factors. The notation $\nabla u, \nabla^2 u, \dots, \nabla^k u$ collectively signifies all partial derivatives of $u$ up to order $k$.

For example, consider the time-dependent diffusion equation in one spatial dimension:
\begin{equation}
\frac{\partial u}{\partial t} = D \frac{\partial^2 u}{\partial x^2}.
\end{equation}
Here, $u(x, t)$ is the unknown function representing concentration or temperature, $x$ and $t$ are independent variables (space and time), and $D$ is the diffusion coefficient, which serves as the parameter $\mathbf{p}$. When studying diffusion in different materials, $D$ will assume various values (e.g., $D_1, D_2, \dots, D_m$). Each such value defines a distinct PDE within the diffusion equation family, all sharing the same core mathematical structure. Solving a PDE family thus entails finding solutions for $u(x, t; D)$ across a specified range of $D$ values.

\section{detailed experimental setup} \label{sec:es}

\subsection{Datasets Details}
\subsubsection{1D Advection Equation}
The 1D Adevction equation describes the transport of a conserved quantity along the spatial domain under a constant velocity. 
The equation is formulated as:
\begin{equation}
    \frac{\partial u}{\partial t} + \beta \frac{\partial u}{\partial x} = 0, \label{eq:1d_advection}
\end{equation}
where $\beta$ denotes the constant advection velocity. The system is subject to an initial condition $u(x, t = 0) = u_0(x)$. Notably, this equation admits an analytical solution: $u(x,t) = u_0(x - \beta t)$.

For our implementation, we construct the initial condition through a superposition of $N$ sinusoidal waves:
\begin{equation}
u_0(x) = \sum_{i=1}^{N} A_i \sin(k_i x + \phi_i), \label{eq:1d_advection_ic}
\end{equation}
where $k_i = \frac{2\pi n_i}{L_x}$, with $n_i$ being random integers sampled from the interval $[1, n_{max}]$. Here, $N$ represents the number of sinusoidal components, $L_x$ denotes the computational domain size, $A_i$ are random float numbers uniformly chosen in $[0, 1]$, and $\phi_i$ represents randomly selected phases from the interval $[0, 2\pi]$.

\begin{figure*}[t]
    \centerline{\includegraphics[scale=0.5]{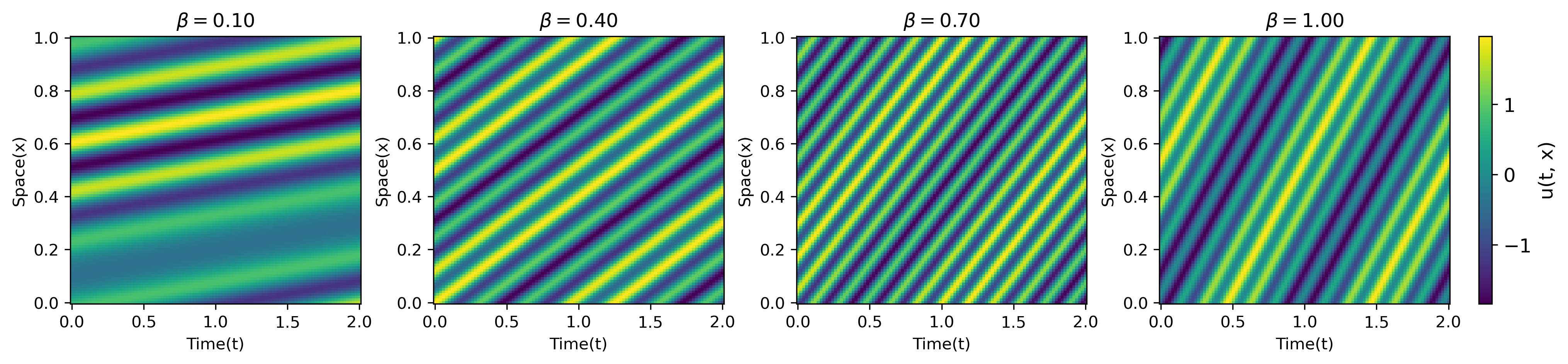}}
    \caption{Visualization of solutions for the 1D Advection equation under different parameters.}
    \label{fig:adv}
\end{figure*}

In our experimental setup, we configure $N = 2$ and $n_{max} = 8$. The dataset is generated directly from the analytical solution by sampling points in the spatial domain $x \in [0, 1]$ and temporal domain $t \in [0, 2]$. We generate $N_{\text{tasks}} = 4$ distinct tasks, where each task $j$ is assigned a specific advection velocity $\beta$ from the set $\{0.1,~0.4,~0.7,~1.0\}$, which is linearly spaced in the range $[0.1,~1.0]$. Fig.~\ref{fig:adv} provides a visualization of the solutions for the 1D Advection equation under different parameter settings.

\subsubsection{1D Burgers' Equation}

The 1D Burgers' equation combines nonlinear convection and viscous dissipation, and serves as a canonical model for studying turbulence and shock formation in fluid dynamics. 
The equation is formulated as:
\begin{equation}
    \frac{\partial u}{\partial t} + \frac{\partial}{\partial x} \left( \frac{u^2}{2} \right) = \frac{\nu}{\pi} \frac{\partial^2 u}{\partial x^2}, \label{eq:burgers}
\end{equation}
where $\nu$ denotes the kinematic viscosity, and $u(x,t)$ is the velocity field. The system is subject to periodic boundary conditions and an initial condition $u(x, t = 0) = u_0(x)$. Unlike the advection equation, this equation does not have a simple general analytical solution for an arbitrary $u_0(x)$. Therefore, the dataset is generated using a numerical solver. We employ a finite difference method on a grid with $M=100$ spatial points and $T=1000$ time steps to solve the equation.

\begin{figure*}[t]
    \centerline{\includegraphics[scale=0.5]{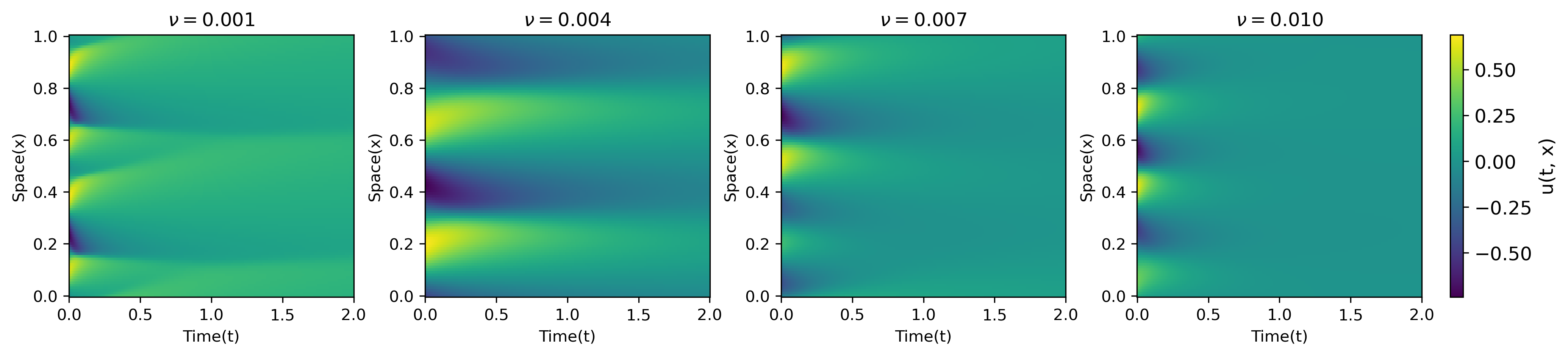}}
    \caption{Visualization of solutions for the 1D Burgers' equation under different parameters.}
    \label{fig:burgers}
\end{figure*}

For our implementation, we adopt the same initial condition as in the 1D advection case. The dataset is generated by first solving the PDE numerically over the spatial domain $x \in [0, 1]$ and temporal domain $t \in [0, 2]$. Then, $1100$ data points are randomly sampled from the resulting solution grid. We generate $N_{\text{tasks}} = 4$ distinct tasks, where each task $j$ is assigned a specific kinematic viscosity $\nu$ from the set $\{0.001,~0.004,~0.007,~0.01\}$. Fig.~\ref{fig:burgers} provides a visualization of the solutions for the 1D Burgers' equation under different parameter settings.

\subsubsection{1D Advection-Diffusion Equation}

The 1D Advection-Diffusion equation characterizes the interplay between advective transport and diffusive spreading of a scalar field. 
The equation is formulated as:
\begin{equation}
    \frac{\partial u}{\partial t} + \beta \frac{\partial u}{\partial x} = \alpha \frac{\partial^2 u}{\partial x^2}, \label{eq:advection-diffusion}
\end{equation}
where $\beta$ denotes the advection velocity, $\alpha$ represents the diffusion coefficient, and $u(x,t)$ is the conserved quantity. The system is subject to an initial condition $u(x, t = 0) = u_0(x)$. This equation does not have a simple general analytical solution for an arbitrary $u_0(x)$, so the dataset is generated using a numerical solver. We employ the Crank-Nicolson method, a finite difference scheme, on a grid with $N_x = 100$ spatial points and $N_t = 100$ time steps to solve the equation.

\begin{figure*}[tp]
    \centerline{\includegraphics[scale=0.5]{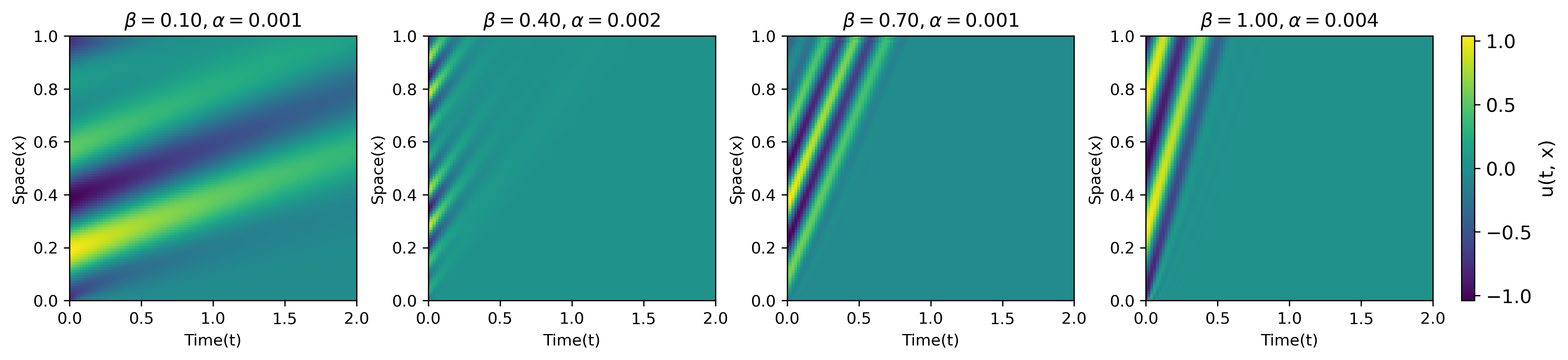}}
    \caption{Visualization of solutions for the 1D Advection-Diffusion equation under different parameters.}
    \label{fig:adv_diff}
\end{figure*}

For our implementation, we adopt the same initial condition as in the 1D advection case. The dataset is generated by first solving the PDE numerically over the spatial domain $x \in [0, 1]$ and temporal domain $t \in [0, 2]$. Then, $1100$ data points are randomly sampled from the resulting solution grid. We generate $N_{\text{tasks}} = 4$ distinct tasks. Each task $j$ is assigned a specific advection velocity $\beta$ from the set $\{0.1,~0.4,~0.7,~1.0\}$, which is linearly spaced in the range $[0.1, 1.0]$. Concurrently, each task is also assigned a unique, randomly sampled diffusion coefficient $\alpha$ from the interval $\mathcal{U}(0.001, 0.005)$. Fig.~\ref{fig:adv_diff} provides a visualization of the solutions for the 1D Advection-Diffusion equation under different parameter settings.

\begin{figure*}[t]
    \centerline{\includegraphics[scale=0.38]{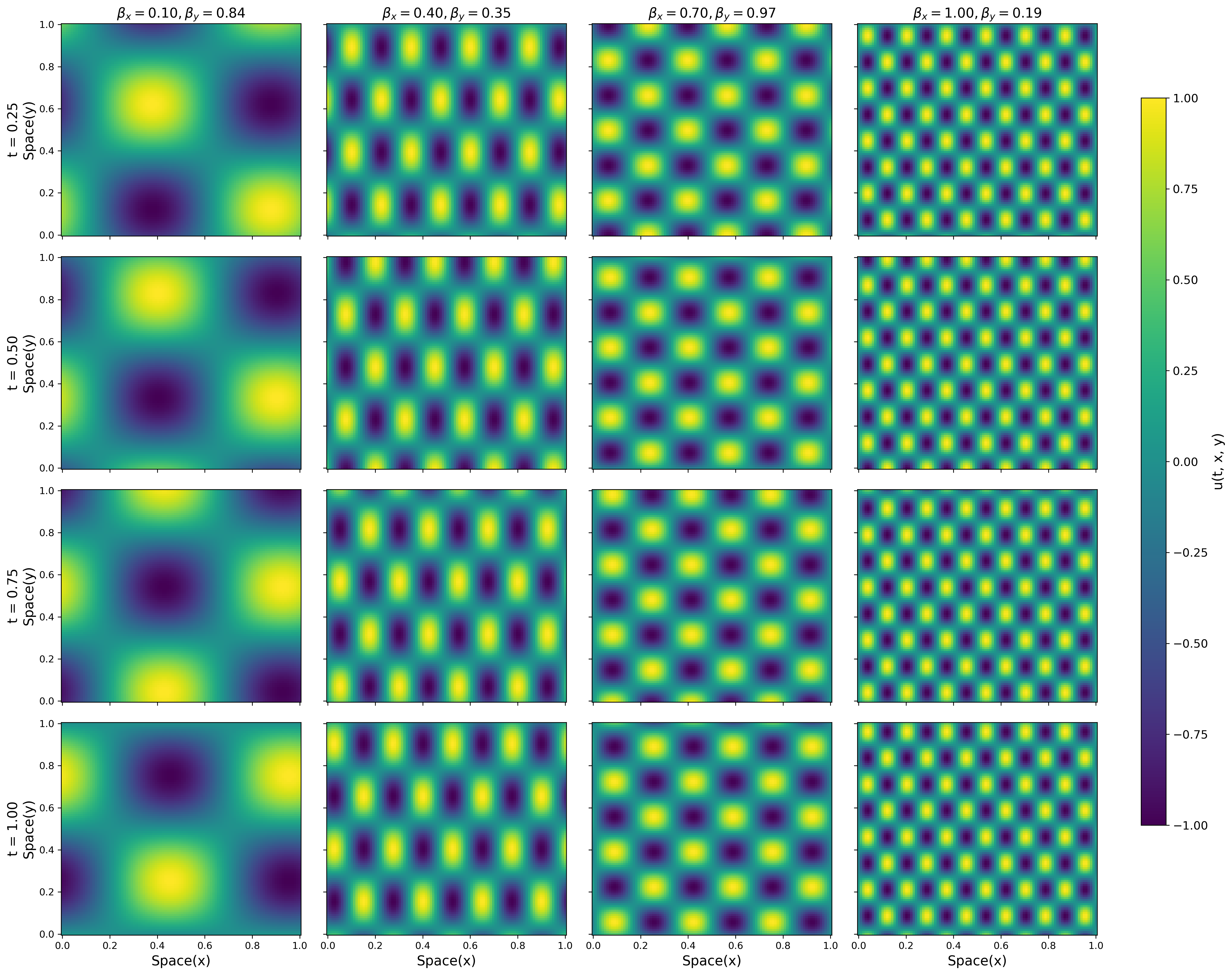}}
    \caption{Visualization of solutions for the 2D Advection equation under different parameters.}
    \label{fig:adv2d}
\end{figure*}

\subsubsection{2D Advection Equation}

The 2D Advection equation describes transport along a planar flow field and tests solver performance under multi-dimensional spatial interactions. 
The equation is formulated as:
\begin{equation}
    \frac{\partial u}{\partial t} + \beta_x \frac{\partial u}{\partial x} + \beta_y \frac{\partial u}{\partial y} = 0, \label{eq:2d_advection}
\end{equation}
where $\beta_x$, $\beta_y$ denote the advection velocities in the $x$ and $y$ directions, respectively. The system is subject to an initial condition $u(x, y, t = 0) = u_0(x, y)$. Notably, this equation admits an analytical solution: $u(t, x, y) = u_0(x - \beta_x t, y - \beta_y t)$.

For our implementation, we construct the initial condition as a product of two sinusoidal waves:
\begin{equation}
u_0(x, y) = A_1 \sin(k_1 x + \phi_1) \cdot A_2 \sin(k_2 y + \phi_2), \label{eq:2d_advection_ic}
\end{equation}
where $k_i = 2\pi n_i$, with $n_i$ being random integers sampled from the interval $[1, n_{max}]$. This corresponds to a computational domain size of $L_x = 1$ and $L_y = 1$. Here, the amplitudes $A_1$ and $A_2$ are fixed at $1.0$, and $\phi_i$ represents randomly selected phases from the interval $[0, 2\pi]$.

In our experimental setup, we configure $n_{max} = 8$. The dataset is generated directly from the analytical solution by sampling $1100$ points in the spatial domain $x, y \in [0, 1]$ and temporal domain $t \in [0, 1]$. We generate $N_{\text{tasks}} = 4$ distinct tasks. Each task $j$ is assigned a specific advection velocity $\beta_{x}$ from the set $\{0.1,~0.4,~0.7,~1.0\}$, which is linearly spaced in the range $[0.1, 1.0]$. Concurrently, each task is also assigned a unique, randomly sampled advection velocity $\beta_{y}$ from the interval $\mathcal{U}(0.1, 1.0)$. Fig.~\ref{fig:adv2d} provides a visualization of the solutions for the 2D Advection equation under different parameter settings.

\begin{figure*}[tp]
    \centerline{\includegraphics[scale=0.35]{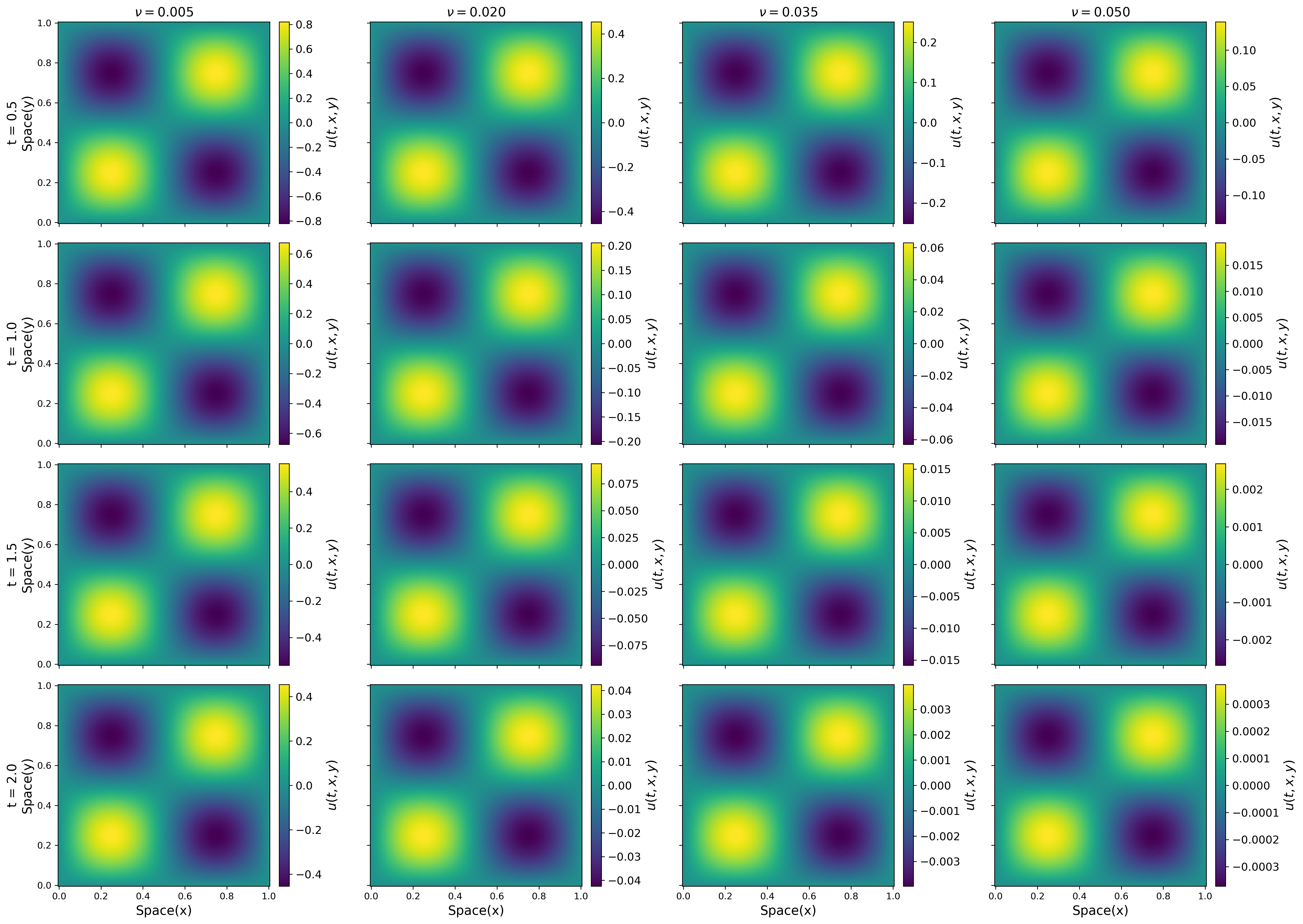}}
    \caption{Visualization of solutions for the 2D Navier-Stokes equation under different parameters.}
    \label{fig:ns2d}
\end{figure*}

\subsubsection{2D Navier-Stokes Equation}

The 2D Navier-Stokes equation models 2D incompressible vorticity evolution, combining nonlinear convection and viscous diffusion to study vortex dynamics and turbulence. 
The equation is formulated as:
\begin{equation}
 \frac{\partial \omega}{\partial t} + \mathbf{u} \cdot \nabla \omega = \nu \Delta \omega, \label{eq:ns_vorticity}
\end{equation}
where $\omega$ is the vorticity, $\mathbf{u}=(u,v)$ is the velocity field, and $\nu$ is the kinematic viscosity. The system is subject to periodic boundary conditions and an initial condition $\omega(x, y, t = 0) = \omega_0(x, y)$. For this specific case, known as the Taylor-Green vortex, the non-linear advection term $\mathbf{u} \cdot \nabla \omega$ simplifies to zero, allowing for a known analytical solution.

For our implementation, we use the specific initial condition for the Taylor-Green vortex: 
\begin{equation} 
\omega_0(x, y) = \sin(2\pi x) \sin(2\pi y). \label{eq:ns_ic} 
\end{equation} 
This initial condition leads to the exact analytical solution: \begin{equation} 
\omega(t, x, y) = \sin(2\pi x) \sin(2\pi y) \exp(-8\pi^2 \nu t). \label{eq:ns_solution} 
\end{equation}

In our experimental setup, the dataset is generated directly from this analytical solution by sampling $1100$ points in the spatial domain $x, y \in [0, 1]$ and temporal domain $t \in [0, 2]$. We generate $N_{\text{tasks}} = 4$ distinct tasks, where each task $j$ is assigned a specific kinematic viscosity $\nu$ from the set $\{0.005,~0.02,~0.035,~0.05\}$, which is linearly spaced in the range $[0.005, 0.05]$. Fig.~\ref{fig:ns2d} provides a visualization of the solutions for the 2D Navier-Stokes equation under different parameter settings.

\begin{figure*}[tp]
    \centerline{\includegraphics[scale=0.3]{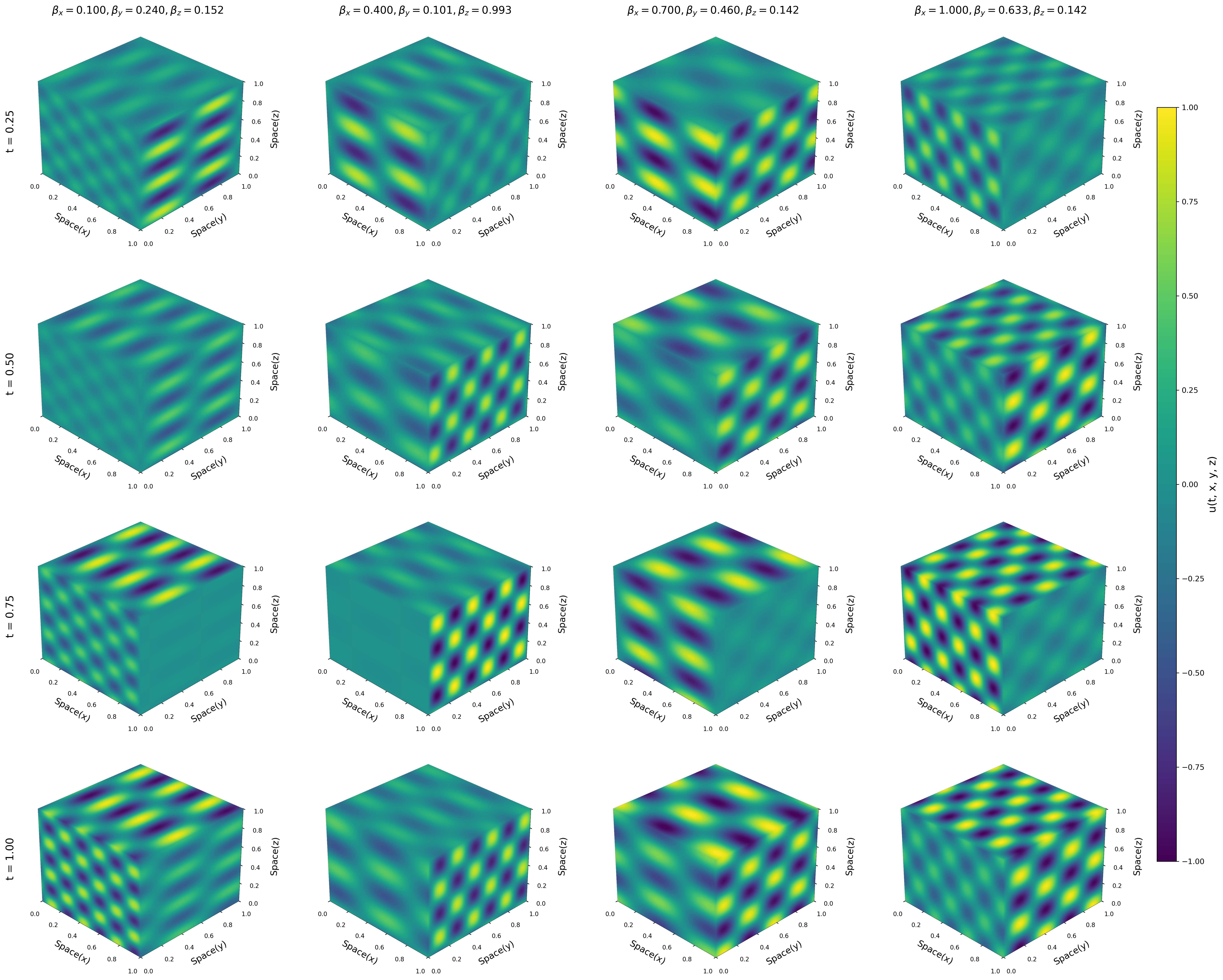}}
    \caption{Visualization of solutions for the 3D Advection equation under different parameters.}
    \label{fig:adv3d}
\end{figure*}

\subsubsection{3D Advection Equation}

The 3D Advection equation evaluates scalability by modeling transport in volumetric flow fields. 
The equation is formulated as:
\begin{equation}
    \frac{\partial u}{\partial t} + \beta_x \frac{\partial u}{\partial x} + \beta_y \frac{\partial u}{\partial y} + \beta_z \frac{\partial u}{\partial z} = 0, \label{eq:3d_advection}
\end{equation}
where $\beta_x$, $\beta_y$, $\beta_z$ denote the advection velocities in the $x$, $y$, and $z$ directions, respectively. The system is subject to an initial condition $u(x, y, z, t = 0) = u_0(x, y, z)$. Notably, this equation admits an analytical solution: $u(t, x, y, z) = u_0(x - \beta_x t, y - \beta_y t, z - \beta_z t)$.

For our implementation, we construct the initial condition as a product of three sinusoidal waves:
\begin{equation}
u_0(x, y, z) = A_1 \sin(k_1 x + \phi_1) \cdot A_2 \sin(k_2 y + \phi_2) \cdot A_3 \sin(k_3 z + \phi_3), \label{eq:3d_advection_ic}
\end{equation}
where $k_i = 2\pi n_i$, with $n_i$ being random integers sampled from the interval $[1, n_{max}]$. This corresponds to a computational domain size of $L_x = L_y = L_z = 1$. Here, the amplitudes $A_1, A_2, A_3$ are fixed at $1.0$, and $\phi_i$ represents randomly selected phases from the interval $[0, 2\pi]$.

In our experimental setup, we configure $n_{max} = 8$. The dataset is generated directly from the analytical solution by sampling $1100$ points in the spatial domain $x, y, z \in [0, 1]$ and temporal domain $t \in [0, 1]$. We generate $N_{\text{tasks}} = 4$ distinct tasks. Each task $j$ is assigned a specific advection velocity $\beta_{x}$ from the set $\{0.1,~0.4,~0.7,~1.0\}$, which is linearly spaced in the range $[0.1, 1.0]$. Concurrently, each task is also assigned a unique, randomly sampled advection velocity $\beta_{y}$ and $\beta_{z}$ from the interval $\mathcal{U}(0.1, 1.0)$. Fig.~\ref{fig:adv3d} provides a visualization of the solutions for the 3D Advection equation under different parameter settings.
        
\subsection{Baselines Details}
In this section, we provide detailed descriptions of the five baseline methods used for comparison in our experiments:
\begin{itemize}
    \item PhySO \cite{tenachi2023deep}: A physics-constrained symbolic regression framework combining neural network-guided search with physical priors and dimensional constraints, excelling in interpretable PDE discovery.
    
    \item DSR \cite{petersen2021deep}: A deep learning-based symbolic regression approach using recurrent neural networks and risk-seeking policy gradients to efficiently recover mathematical expressions from noisy or noiseless data.
    
    \item SP-GPSR \cite{cao2023genetic}: A genetic programming symbolic regression method employing simplification-pruning operators and multi-objective fitness evaluation, yielding compact analytical solutions for complex differential equations.
    
    \item GNOT \cite{hao2023gnot}: A transformer-based neural operator encoding PDEs via graph structures, supporting multi-input and irregular meshes for scalable multi-task PDE learning.
    
    \item Geo-FNO \cite{li2023fourier}: A Fourier neural operator framework that incorporates learned domain deformations to map physical domains onto latent spaces, enabling the solution of PDEs on arbitrary geometries.
    
\end{itemize}

\subsection{Hyper-parameters Details}
For NMIPS, the experimental hyperparameters were set as follows: population size $N=50$, random mating probability $rmp=0.2$, mutation probability $0.002$, number of generations $100$, and maximum number of evaluations $5000$. Relevant parameters in baseline methods, such as population or iteration counts, were kept consistent to ensure fair comparisons. Regarding the symbolic library, for the 1D, 2D, and 3D advection equations, the candidate symbol library consists of $\{+, -, \times, \sin\}$, excluding PDE parameters. For the other three equations, the library is extended to $\{+, -, \times, \sin, \exp, \log\}$ to accommodate the increased functional complexity.

\section{more experimental results}

\subsection{More Results on the 1D Advection Equation}
\begin{figure*}[t]
    \centerline{\includegraphics[scale=0.38]{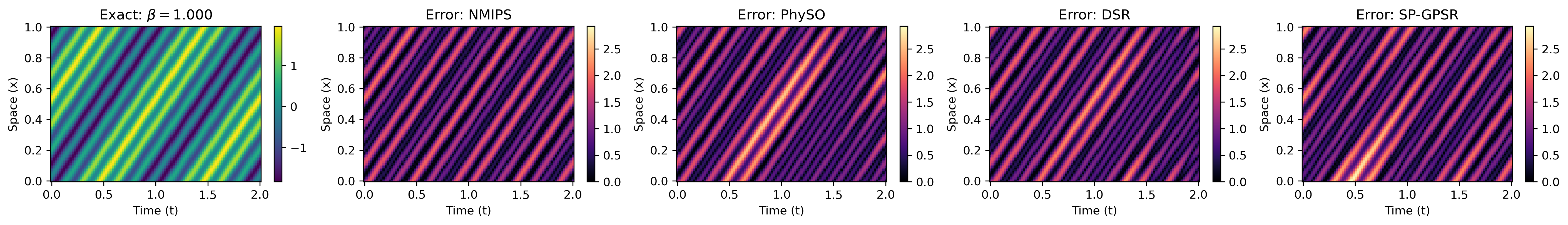}}
    \caption{A spatial-temporal error analysis of different methods for the 1D Advection equation.}
    \label{fig:adv_error}
\end{figure*}

To provide a more granular assessment of model fidelity, Fig.~\ref{fig:adv_error} visualizes the spatial-temporal error distributions for the specific case of $\beta=1.000$. The ``Exact" panel displays the characteristic traveling wave patterns of the advection equation. As shown in the error heatmaps, NMIPS yields a predominantly dark field, indicating minimal pointwise error throughout the simulation domain. In contrast, the error maps for baselines such as PhySO and SP-GPSR exhibit prominent, bright bands of high error aligned with the wave propagation. These artifacts suggest that the baseline methods struggle to perfectly capture the phase velocity or amplitude of the traveling waves, whereas NMIPS maintains high fidelity to the exact dynamics over time.

\subsection{More Results on the 1D Advection-Diffusion Equation}
\begin{figure*}[t]
    \centerline{\includegraphics[scale=0.38]{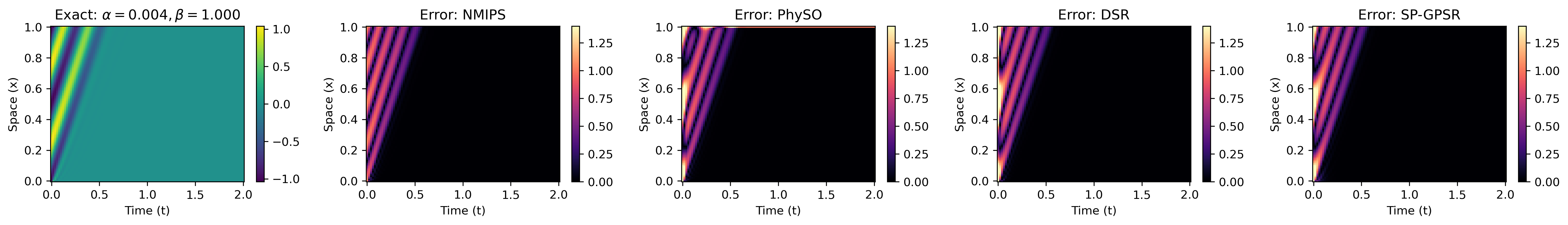}}
    \caption{A spatial-temporal error analysis of different methods for the 1D Advection-Diffusion equation.}
    \label{fig:adv_diff_error}
\end{figure*}

Qualitative results in Fig.~\ref{fig:adv_diff_error} further illustrate NMIPS's robustness. The error heatmaps demonstrate that NMIPS maintains minimal pointwise error (indicated by the dark field) even in advection-dominated regions. In contrast, baseline methods like PhySO and SP-GPSR exhibit distinct bright error bands aligned with the direction of wave propagation, suggesting a limitation in capturing the precise phase and amplitude of the steep gradients. This confirms that NMIPS possesses strong generalization capabilities, effectively balancing the dynamics of advection and diffusion.

\begin{table*}[t!]
    \centering
    \setlength{\tabcolsep}{5pt}
    \caption{MSE of different methods on 1D Advection-Diffusion equation}
    \begin{tabular}{cccccccccccc}
        \toprule
        \multirow{2}*{\textit{\textbf{Num}}} & \multirow{2}*{\textbf{$\alpha$}} & \multirow{2}*{\textbf{$\beta$}} & \multicolumn{4}{c}{\textbf{NMIPS}} & \multirow{2}*{\textbf{PhySO}} & \multirow{2}*{\textbf{DSR}} & \multirow{2}*{\textbf{SP-GPSR}} & \multirow{2}*{\textbf{GNOT}} & \multirow{2}*{\textbf{Geo-FNO}} \\
         & & & 0 & 1 & 2 & 3 & & & & & \\
        \midrule
        0 & 0.100  & 0.001  & / & 2.61E-01 & 2.78E-01 & 2.60E-01 & 2.52E-01 & 2.30E-01 & 2.70E-01 & 5.44E-01 & 3.15E-01 \\
        1 & 0.400  & 0.002  & 1.08E-02 & / & 1.08E-02 & 1.08E-02 & 2.62E-02 & 1.91E-02 & 1.15E-02 & 3.11E-01 & 1.89E-02 \\
        2 & 0.700  & 0.001  & 1.47E-01 & 1.48E-01 & / & 1.49E-01 & 1.71E-01 & 1.63E-01 & 1.47E-01 & 3.35E-01 & 1.56E-01 \\
        3 & 1.000  & 0.004  & 5.13E-02 & 5.13E-02 & 5.11E-02 & / & 8.01E-02 & 9.11E-02 & 7.18E-02 & 3.52E-01 & 6.58E-02 \\
        \midrule
        \multicolumn{3}{c}{\textbf{Avg}} & \multicolumn{4}{c}{\textbf{1.19E-01}}  & 1.32E-01 & 1.26E-01 & 1.25E-01 & 3.85E-01 & 1.39E-01 \\
        \bottomrule
    \end{tabular}
    \label{base:adv_diff_1d_1}
\end{table*}

\subsection{More Results on the 2D Advection Equation}
\begin{figure*}[t]
    \centerline{\includegraphics[scale=0.39]{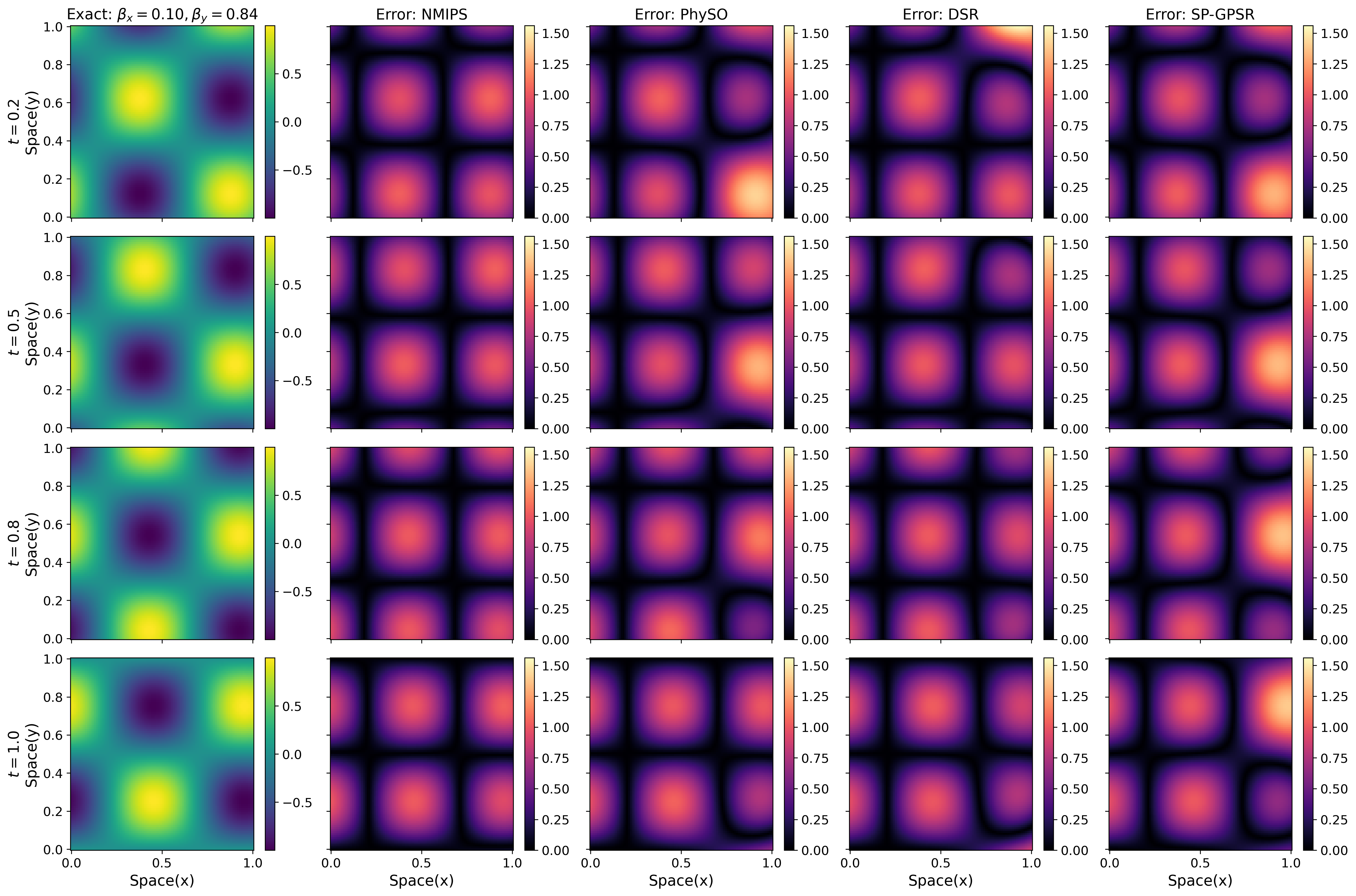}}
    \caption{A spatial-temporal error analysis of different methods for the 2D Advection equation.}
    \label{fig:adv2d_error}
\end{figure*}

Qualitative results in Fig.~\ref{fig:adv2d_error} further highlight the precision of NMIPS on the 2D Advection equation. The error heatmaps reveal that NMIPS achieves a uniformly low error distribution (predominantly dark field) across the entire spatial domain over time. In contrast, baseline methods such as PhySO and SP-GPSR display distinct localized regions of high error (bright spots), particularly concentrated near the domain corners or peak value regions. This discrepancy suggests that while competing methods may capture the general wave pattern, they struggle to maintain accuracy at the extrema, whereas NMIPS accurately reconstructs the full spatial field with high fidelity.

\subsection{More Results on the 2D Navier-Stokes Equation}
The source of NMIPS's accuracy is further elucidated by the symbolic expressions in Table \ref{tab:ns_sol}. While NMIPS discovers a structured, interpretable expression involving linear and trigonometric terms ($u \propto (x - c) \sin(y)$), baseline methods converge on highly convoluted forms involving nested logarithms and exponentials (e.g., $\sin(\log(x))$ in DSR). This contrast indicates that NMIPS effectively balances numerical precision with physical parsimony, avoiding the overfitting artifacts commonly observed in competing approaches.

\begin{figure*}[t!]
    \centerline{\includegraphics[scale=0.39]{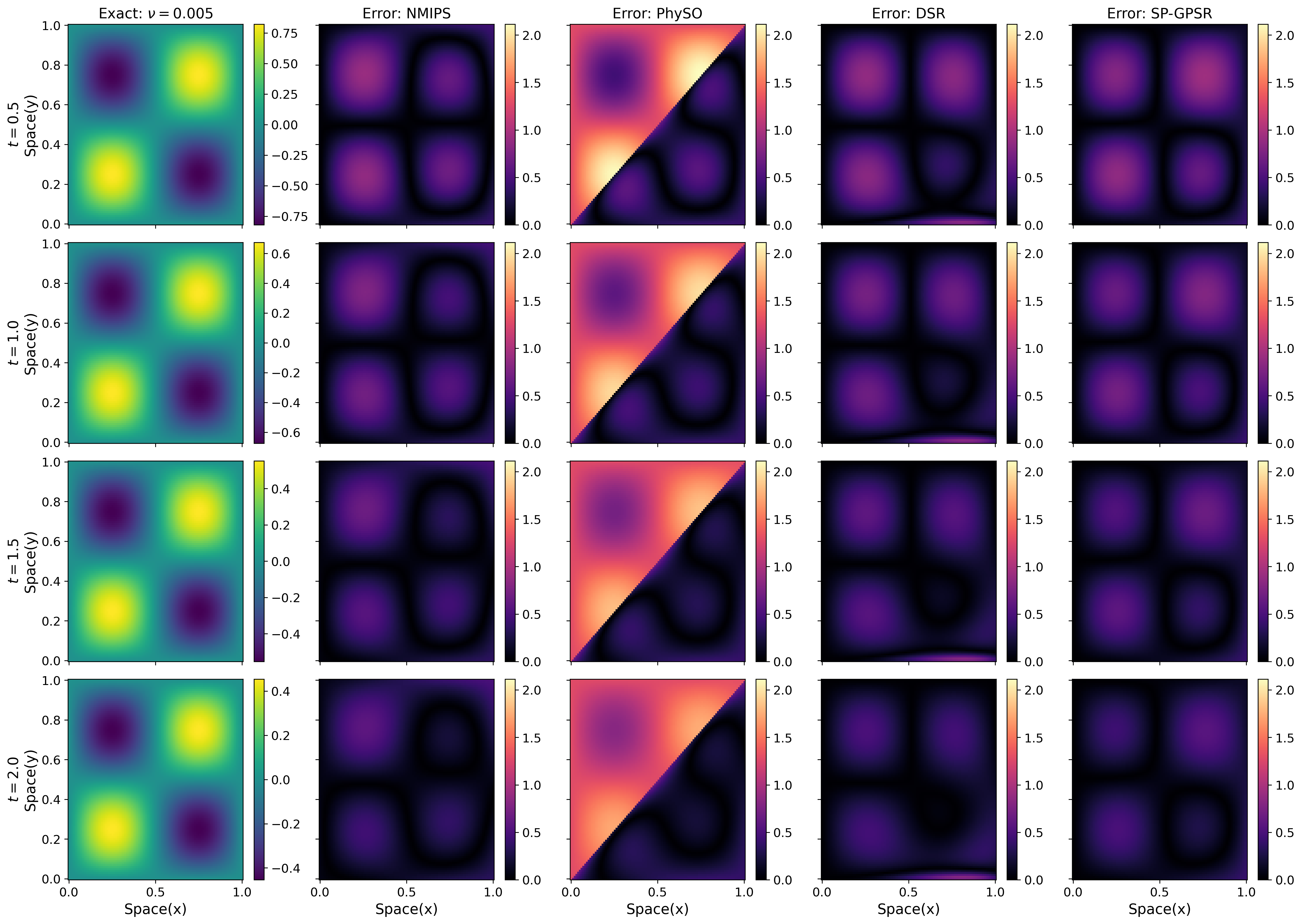}}
    \caption{A spatial-temporal error analysis of different methods for the 2D Navier-Stokes equation.}
    \label{fig:ns_error}
\end{figure*}

This structural superiority directly translates into the robust predictive performance visualized in Fig.~\ref{fig:ns_error}. The error heatmaps provide compelling evidence of NMIPS's capability in handling complex fluid dynamics. NMIPS maintains a consistently dark error profile, indicating minimal deviation from the exact solution even under non-linear conditions. In contrast, the incorrect symbolic forms of the baselines manifest as severe failure modes: PhySO exhibits a sharp diagonal artifact and a massive region of high error (bright triangular area) in the lower half of the domain, while DSR and SP-GPSR show visible distributed error patterns. This confirms that by successfully identifying the correct governing physical laws, NMIPS avoids the structural instabilities that lead to significant deviations in baseline models.

\begin{table*}[t]
    \centering
    \setlength{\tabcolsep}{5pt}
    \caption{MSE of different methods on 2D Navier-Stokes equation}
    \begin{tabular}{cccccccccccc}
        \toprule
        \multirow{2}*{\textit{\textbf{Num}}} & \multirow{2}*{\textbf{$\nu$}} & \multicolumn{4}{c}{\textbf{NMIPS}} & \multirow{2}*{\textbf{PhySO}} & \multirow{2}*{\textbf{DSR}} & \multirow{2}*{\textbf{SP-GPSR}} & \multirow{2}*{\textbf{GNOT}} & \multirow{2}*{\textbf{Geo-FNO}} \\
        & & 0 & 1 & 2 & 3 & & & & & \\
        \midrule
        0 & 0.005  & / & 1.19E-01 & 1.18E-01 & 1.31E-01 & 9.86E-02 & 1.16E-01 & 1.30E-01 & 7.58E-02 & 1.74E-01 \\
        1 & 0.020  & 4.36E-02 & / & 4.47E-02 & 5.04E-02 & 1.14E-01 & 4.50E-02 & 4.45E-02 & 1.07E-01 & 4.74E-02 \\
        2 & 0.035  & 2.36E-02 & 2.99E-02 & / & 2.10E-02 & 2.06E-02 & 3.52E-02 & 2.19E-02 & 1.12E-01 & 4.12E-02 \\
        3 & 0.050  & 1.40E-02 & 2.09E-02 & 1.44E-02 & / & 2.09E-02 & 1.52E-02 & 1.44E-02 & 1.10E-01 & 2.24E-02 \\
        \midrule
        \multicolumn{2}{c}{\textbf{Avg}} & \multicolumn{4}{c}{\textbf{5.25E-02}} & 6.36E-02 & 5.29E-02 & 5.26E-02 & 1.01E-01 & 7.13E-02 \\
        \bottomrule
    \end{tabular}
    \label{base:ns2d}
\end{table*}

\begin{table}[t!]
  \centering
  \caption{Discovered expressions for the 2D Navier-Stokes equation by different symbolic regression methods}
  \begin{tabular}{cc}
    \toprule
    \textbf{Method} & \textbf{Expression} \\
    \midrule
    NMIPS & $u = (x - 0.493)\,\sin(y) + 0.037\,t$ \\
    PhySO & $\displaystyle u = -\sin\!\big( 0.332\,(\log(y - x) - y\,(y - 0.589)) + 0.195 \big) - 0.332$ \\
    DSR & $u = (y - x)\,\sin\!\big( y\,e^{y} \big)\,\sin\!\big( \log(x) \big)$ \\
    SP-GPSR & $u = -e^{y - x}\,\sin(0.150\,y)$ \\
    \bottomrule
  \end{tabular}
  \label{tab:ns_sol}
\end{table}

\subsection{More Results on the 3D Advection Equation}
Quantitative superiority is directly explained by the discovered expressions in Table \ref{tab:adv3d_sol}. NMIPS identifies a clean, physically plausible polynomial form that correctly captures the linear advective relationship between spatial and temporal coordinates. In contrast, competing methods fail to uncover the true governing structure: PhySO and DSR converge on complex, non-intuitive trigonometric functions, while SP-GPSR yields incorrect interaction terms. The failure of baseline methods to identify the correct symbolic form leads to their higher prediction errors, whereas NMIPS effectively balances numerical stability with the recovery of simple, interpretable solutions in high-dimensional spaces.

\begin{table}[t]
  \centering
  \caption{Discovered expressions for the 3D Advection equation by different symbolic regression methods}
  \begin{tabular}{cc}
    \toprule
    \textbf{Method} & \textbf{Expression} \\
    \midrule
    NMIPS & $u = 0.0000352 \cdot y z^2 (2x - t)$ \\
    PhySO & $u = \sin\!\left( z \left( 6.244\,x\,(z - 0.558)^2 - 0.295 \right) \right)$ \\
    DSR & $u = z^3 \sin\!\left( z(x + 1) - x t - 1 \right)$ \\
    SP-GPSR & $u = 1.217\,z\,(t - x)\,(y - z)$ \\
    \bottomrule
  \end{tabular}
  \label{tab:adv3d_sol}
\end{table}

\begin{figure}[t]
    \centering
    \includegraphics[scale=0.5]{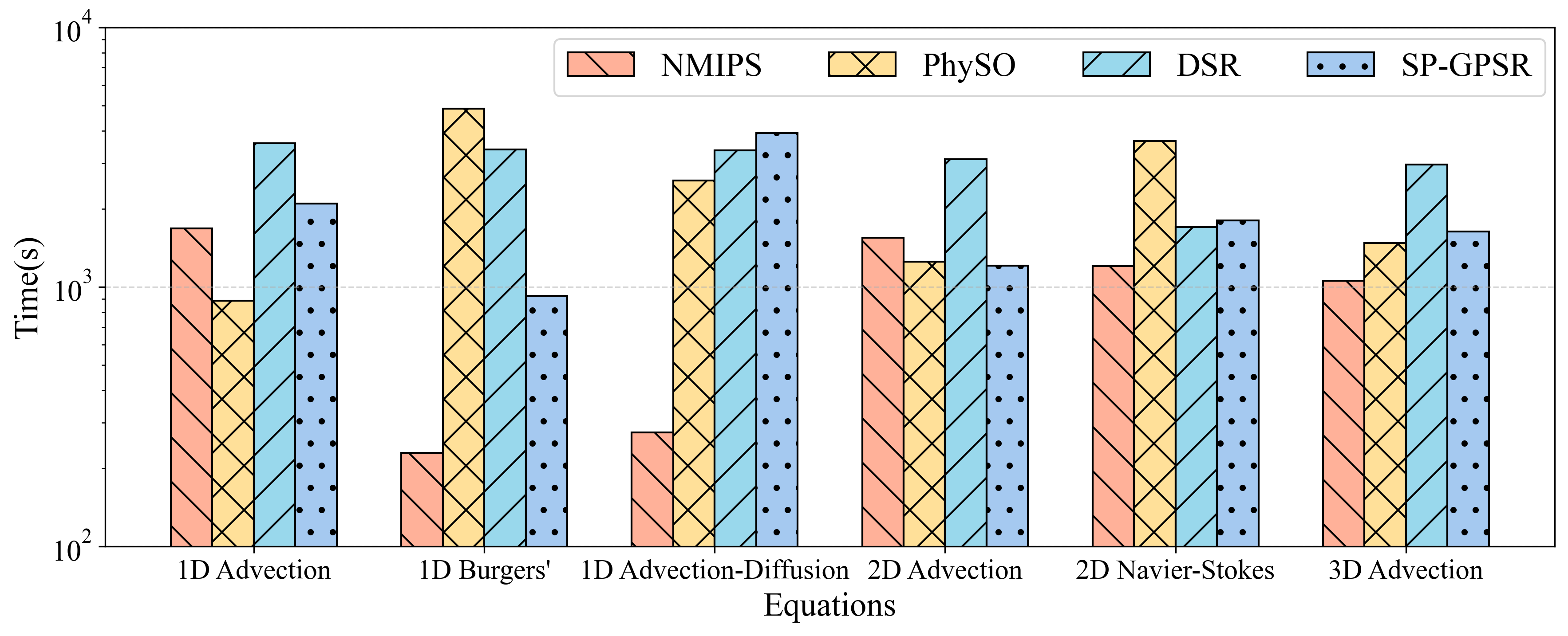}
    \caption{Runtime comparison of different methods on different equations.}
    \label{fig:runtime}
\end{figure}

\section{Scientific Insight}

\begin{table*}[t!]
    \centering
    \caption{Summary of invariant symbolic skeletons discovered by NMIPS across varying PDE families}
    \label{tab:scientific_discovery}
    \begin{tabular}{cccc}
    \toprule
    \textbf{PDE Family} & \textbf{Dimensions} & \textbf{Dominant Physics} & \textbf{Key Invariant Skeleton} \\ 
    \midrule
    \textbf{Advection} & 1D & Linear Transport & $x - \beta t$ \\
    \textbf{Advection} & 2D / 3D & Multi-dim Transport & $x_i - \beta_i t$ \\
    \textbf{Burgers'} & 1D & Nonlinearity + Shock & $\ln(\dots)$ or $x-t$ \\
    \textbf{Advection-Diffusion} & 1D & Transport + Dissipation & $\exp(f(x) - c \cdot t)$ \\
    \textbf{Navier-Stokes} & 2D & Vorticity + Viscosity & $\exp(-ct) \sin(x) \sin(y)$ \\ 
    \bottomrule
    \end{tabular}
\end{table*}

Moving beyond numerical accuracy, a critical capability of AI-driven PDE solvers is the autonomous discovery of the underlying ``scientific skeleton'' of physical systems. We analyzed the converged expression trees discovered by NMIPS across six diverse PDE families. The results reveal that our method successfully captures invariant physical mechanisms, ranging from characteristic lines in hyperbolic systems to the separation of variables in complex fluid dynamics.

For hyperbolic transport problems, the fundamental physical invariant is the concept of characteristic lines. Specifically, in the 1D Advection tasks with varying velocities ($\beta$), NMIPS consistently converged on linear substructures of the form $(x - \beta t)$. This indicates the algorithm successfully ``rediscovered'' that the solution is a traveling wave that propagates without deformation. Crucially, this capability scales to higher dimensions. In the 2D and 3D Advection tasks, the solver demonstrated dimensional decoupling. For instance, in 3D flows, it isolated specific transport axes, recovering terms like $(z - t)$ or $(y - \beta t)$. This confirms that NMIPS reconstructs the characteristic hyperplanes governing volumetric transport rather than merely memorizing spatial patterns.

In systems where multiple physical mechanisms compete, NMIPS demonstrated the ability to disentangle these effects symbolically. For the 1D Advection-Diffusion equation, the solver identified a composite structure: a traveling wave argument $(x - ct)$ modulated by an exponential decay envelope $\exp(-kt)$. This signifies a physical understanding that diffusion acts as an irreversible energy dissipation mechanism on top of the convective transport. In the nonlinear Burgers' equation, the solver captured distinct regimes. In low-viscosity tasks, it recovered the inviscid limit approximation $(x - t)$, while in shock-forming regimes, it introduced logarithmic or nested exponential terms to approximate the steep gradients.

Finally, in the 2D Navier-Stokes equation, NMIPS achieved its most significant discovery. The analytical solution is characterized by the separation of spatial and temporal dynamics: $\omega(x,y,t) = \sin(x)\sin(y)e^{-2\nu t}$.
Remarkably, our experimental results show that the solver converged on a symbolic skeleton of the exact form:
\begin{equation}
    u \approx \underbrace{\exp(c_1 \cdot t)}_{\text{Temporal Decay}} \cdot \underbrace{\sin(x) \cdot \sin(c_2 \cdot y)}_{\text{Spatial Mode}}
\end{equation}
This finding confirms that NMIPS can autonomously rediscover the method of separation of variables, correctly modeling the viscous decay of vorticity as a temporal multiplier independent of the spatial flow structure. Table \ref{tab:scientific_discovery} summarizes the dominant symbolic skeletons identified across these PDE families, unequivocally demonstrating that NMIPS acts as a transparent tool for scientific discovery.

\section{Discussion}
\subsection{Negative Transfer and Task Interference}

A critical concern in multitask learning is whether knowledge sharing across tasks may inadvertently degrade performance due to task interference. In NMIPS, such adverse effects are alleviated by performing knowledge transfer at the symbolic structure level rather than through direct parameter sharing. Specifically, candidate solutions are represented as expression trees whose coefficients and fitness evaluations remain task-specific. This design enables the framework to exploit shared mathematical structures among PDE instances while preserving sufficient flexibility to accommodate task-dependent characteristics.

Moreover, the MFO mechanism evaluates each individual strictly with respect to its associated task, and selection is guided by task-aware fitness measures. As a result, solution candidates that are beneficial for one PDE instance but detrimental to others are naturally suppressed during evolution, preventing the accumulation of harmful cross-task biases. This selective and structure-oriented sharing strategy fundamentally differs from conventional neural multitask models that rely on hard or soft parameter sharing.

However, it should be noted that when PDE instances within the same family exhibit substantial differences in initial conditions, boundary conditions, or solution regularity, structure-level transfer may become less effective. In such scenarios, adaptive task grouping or similarity-aware transfer strategies could further enhance robustness, which remains an important direction for future research.

\subsection{Scalability to Higher-Dimensional PDEs}
The experimental results demonstrate that NMIPS remains effective for PDEs up to three spatial dimensions, which already cover the vast majority of physical systems encountered in real-world applications, such as fluid dynamics, transport phenomena, and wave propagation. While PDEs formulated in dimensions higher than three do exist, they typically arise from abstract mathematical constructions or augmented state spaces rather than direct physical modeling.

From a methodological perspective, extending symbolic regression-based PDE solvers to substantially higher dimensions introduces significant challenges. The symbolic search space grows combinatorially with dimensionality, and the evaluation of PDE residuals becomes increasingly expensive, making exhaustive structure exploration impractical. Moreover, higher-dimensional PDEs rarely admit concise analytical solutions, which further reduces the practical utility of symbolic discovery in such settings.

Consequently, NMIPS is primarily targeted at low to moderate dimensional PDEs where analytical structure discovery is both feasible and scientifically meaningful. Addressing scalability beyond this regime may require hybrid approaches that combine symbolic structure learning with neural surrogates or dimensionality reduction techniques.

\bibliography{mybibtex}